\documentclass{article}

\usepackage[final]{neurips_2025}

\usepackage[utf8]{inputenc}
\usepackage[T1]{fontenc}
\usepackage{hyperref}
\usepackage{url}
\usepackage{booktabs}
\usepackage{amsfonts}
\usepackage{amssymb}
\usepackage{amsmath}
\usepackage{amsthm}
\newtheorem{proposition}{Proposition}
\usepackage{nicefrac}
\usepackage{microtype}
\usepackage{graphicx}
\usepackage{xcolor}
\usepackage{multirow}
\usepackage{subcaption}
\usepackage{enumitem}

\title{Why Agent Caching Fails and How to Fix It:\\Structured Intent Canonicalization with Few-Shot Learning}

\author{%
  Abhinaba Basu \\
  \texttt{mail@abhinaba.com}
}

\begin{document}

\maketitle

\begin{abstract}
Personal AI agents executing repetitive tasks through LLM APIs incur substantial cost and latency.
We show that existing caching methods fail on personal agent tasks: GPTCache achieves only 3.3\% cache hit rate on synthetic tasks and 37.9\% accuracy when evaluated as clustering on real benchmarks, while Agentic Plan Caching (APC) achieves 0--12\%.
We identify the root cause: these methods optimize for the wrong property.
Cache effectiveness depends not on classification \emph{accuracy} but on key \emph{consistency} (do equivalent queries always produce the same key?) and \emph{precision} (are cache hits safe to execute?).
We observe that cache-key evaluation reduces to clustering quality evaluation, and apply V-measure decomposition \citep{rosenberg2007vmeasure} to separate these properties on $n{=}8{,}682$ data points across three established benchmarks (MASSIVE, BANKING77, CLINC150) and an additional 1,193 English test examples from NyayaBench v2, our new real-world agentic dataset (8,514 entries, 528 intents $\to$ 20 W5H2 classes, 63 languages).

We introduce W5H2, a structured intent decomposition framework.
Using few-shot contrastive classification (SetFit) with just 8 examples per class, W5H2 achieves 91.1\%$\pm$1.7\% accuracy (mean over 5 seeds) on MASSIVE \citep{fitzgerald2023massive} (8 classes) in ${\sim}$2--5ms---compared to 37.9\% for GPTCache on the same benchmark and 68.8\% for a 20B-parameter LLM baseline at 3,447ms and \$0.09/1K requests.
On NyayaBench v2 (20 real agentic classes---a substantially harder task), SetFit achieves 55.3\%$\pm$1.0\%, rising to 62.6\% with 16 examples per class; zero-shot cross-lingual transfer establishes a baseline across 30 languages (mean 37.7\%, 5 languages above 50\%) with no non-English training data.
On 77-class fine-grained banking intents, SetFit achieves 77.9\% (CI=[76.4\%, 79.4\%]).
We further find that for compound query detection, a smaller NLI model (22M params) achieves 94.4\% while a larger model (280M) achieves 0\%---stronger models over-entail, destroying decision boundaries.

Our five-tier cascade architecture (fingerprint $\to$ BERT $\to$ SetFit $\to$ cheap LLM $\to$ deep agent) handles 85\% of interactions locally at zero cost.
We evaluate across 5 random seeds, compare against fine-tuned BERT (supervised upper bound, 97.3\%) and estimated few-shot LLM baselines (4-shot: ${\sim}$79\%, extrapolated from zero-shot), confirming SetFit's 91.1\%$\pm$1.7\% occupies the favorable few-shot trade-off.
Under our modeled traffic assumptions (Section~\ref{sec:cost}), an individual user would pay \$0.80/month vs.\ \$31.72 for all-LLM---a projected 97.5\% cost reduction.
\end{abstract}

\section{Introduction}
\label{sec:intro}

Personal AI agents---systems like OpenClaw, Manus, and custom assistants---execute user tasks through sequences of LLM API calls.
Each call costs \$0.01--0.10 and takes 2--5 seconds.
Yet most daily interactions follow repetitive patterns with different parameters: ``Check email from Alice'' and ``Check email from Bob'' require the same tool sequence with a single parameter swap.
At 50 requests per day, a single user pays \$26--32/month in API costs alone; service providers report losses of \$10--20K/month at modest scale.

\textbf{Existing caching methods fail for personal agents.}
Exact caching only matches identical inputs and misses paraphrases.
Semantic caching (GPTCache; \citealt{bang2023gptcache}) uses embedding similarity, but ``check email'' and ``send email'' are semantically close yet require completely different tool sequences.
Agentic Plan Caching (APC; \citealt{zhang2025apc}) uses keyword extraction, but personal agent queries are too short (3--8 words) to provide reliable keyword signal.
APC's own paper acknowledges that semantic caching has a ``high rate of false-positive cache hits leading to substantial performance degradation.''
None of these methods work across languages.

\textbf{Our contributions:}
\begin{enumerate}[itemsep=2pt,topsep=2pt]
\item \textbf{W5H2 structured intent decomposition}: action-target cache keys that are language-agnostic and enable partial matching.
No prior work applies structured intent canonicalization to agent plan caching.

\item \textbf{SetFit few-shot for agent caching}: 8 examples per class $\to$ 91.1\%$\pm$1.7\% on MASSIVE (5 seeds), 55.3\% on NyayaBench v2 (20 classes), 84.4\% multilingual, 2.4ms latency.
First application of contrastive few-shot learning to agent plan caching.

\item \textbf{Cache-key = clustering observation}: We observe that evaluating cache-key functions reduces to clustering quality evaluation.
Applying standard V-measure \citep{rosenberg2007vmeasure} decomposes cache quality into precision ($h$) and consistency ($c$) on $n{=}1{,}102$+ data points---replacing ad hoc hit-rate metrics.
Not a new metric, but a new application that enables proper evaluation.

\item \textbf{Rate-distortion analysis}: SetFit operates at or near the information-theoretic rate ($\log_2|\text{keys}| \approx H(\text{Intent})$) on 2/3 benchmarks, suggesting it naturally discovers appropriate compression.

\item \textbf{Counter-intuitive findings}: a 22M-param SetFit model outperforms a 20B-param LLM in zero-shot (91.1\% vs.\ 68.8\%, fully evaluated) and likely with few-shot prompting (${\sim}$79\%, estimated) at $>$700$\times$ lower latency; smaller NLI beats larger for compound detection (94.4\% vs.\ 0\%, preliminary, $n{=}18$); 150 diverse classes beat 77 fine-grained classes; wrong-but-consistent beats right-but-variable.

\item \textbf{Five-tier cascade architecture}: fingerprint $\to$ BERT (supervised) $\to$ SetFit (few-shot) $\to$ cheap LLM $\to$ deep agent, with modeled API cost reduction of 97.5\% under assumed traffic distributions.
The BERT $\to$ SetFit cascade addresses the ``why not just fine-tune BERT?'' question directly: BERT handles known patterns at 97.3\% accuracy; SetFit handles the long tail with 8 examples.
\end{enumerate}

These contributions build on---and diverge from---several lines of prior work, which we survey next.

\begin{figure*}[t]
\centering
\includegraphics[width=\textwidth]{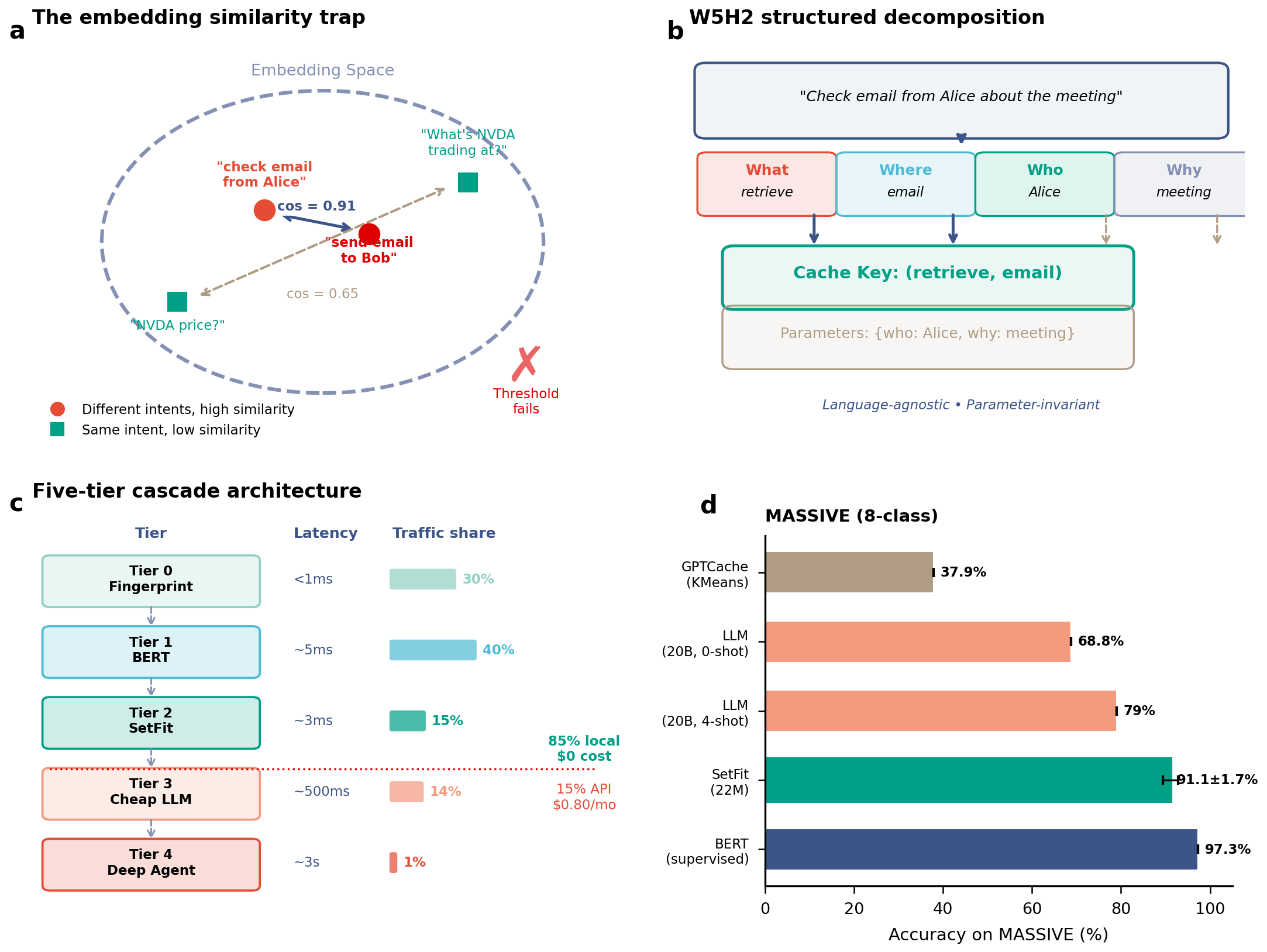}
\caption{Overview of W5H2 structured intent canonicalization for agent caching.
\textbf{(a)}~The embedding similarity trap: semantically similar queries (``check email'' vs.\ ``send email,'' cos$=0.91$) require different tool sequences, while paraphrases of the same intent (cos$=0.65$) fall below typical thresholds.
\textbf{(b)}~W5H2 decomposes queries into structured fields; the (\textsc{What}, \textsc{Where}) pair forms the cache key.
\textbf{(c)}~Five-tier cascade architecture with per-tier latency and traffic share.
\textbf{(d)}~SetFit accuracy on MASSIVE (8-class) across methods: 22M-parameter SetFit (91.1\%) outperforms a 20B-parameter LLM (68.8\%) at $>$700$\times$ lower latency.}
\label{fig:hero}
\end{figure*}

\section{Related Work}
\label{sec:related}

\textbf{LLM Caching.}
GPTCache \citep{bang2023gptcache} caches LLM responses using embedding similarity.
RAGCache \citep{jin2024ragcache} extends caching to retrieval-augmented generation.
\citet{gill2025langcache} fine-tune domain-specific embeddings for semantic caching.
vCache \citep{vcache2025} introduces verified semantic prompt caching with adaptive similarity thresholds.
All of these operate at the embedding similarity level and suffer from the precision--recall tradeoff we identify.

\textbf{Agent Plan Caching.}
APC \citep{zhang2025apc} is closest to our work: it caches tool-use plans keyed by extracted keywords.
Key differences: APC uses keyword extraction (we use structured decomposition + SetFit); APC targets web navigation (we target personal agent tasks); APC is English-only (we evaluate 63 languages); APC does not formalize consistency vs.\ accuracy.

\textbf{Decomposed Caching.}
SemanticALLI \citep{semanticalli2026} independently validates caching reasoning (not just responses) in agentic systems, confirming that structured decomposition of queries improves cache effectiveness.

\textbf{Few-Shot Classification.}
SetFit \citep{tunstall2022setfit} fine-tunes sentence transformers \citep{reimers2019sentencebert} with contrastive learning using 8--16 examples per class.
Prototypical networks \citep{snell2017prototypical} use metric learning for few-shot tasks.
We are the first to apply SetFit to agent plan caching.

\textbf{Query Canonicalization.}
We draw on techniques from databases (prepared-statement fingerprinting), compilers (hash consing), search engines (query normalization), and audio (spectral fingerprinting).
The key insight is that agent intent caching \emph{is} a canonicalization problem---a connection not made in prior NLP/agent work.

\textbf{Intent Recognition.}
Joint intent and slot filling \citep{chen2019jointbert}, DIET classifiers \citep{bunk2020diet}, and the MASSIVE dataset \citep{fitzgerald2023massive} provide the NLU foundation.
We bridge intent recognition with agent plan caching explicitly.

A common thread runs through these efforts: each optimizes one aspect of the caching problem (similarity, keywords, embeddings) without a unifying framework for what makes a cache key \emph{good}.
We address this gap by treating cache-key generation as a canonicalization problem.

\section{W5H2: Intent Canonicalization for Agent Caching}
\label{sec:method}

\subsection{Reframing: Canonicalization, Not Classification}

The standard framing assumes accurate intent \emph{classification}.
Our reframing: what matters is intent \emph{canonicalization}---a stable mapping from surface text to cache keys:
\begin{align}
\text{Classification:}\quad & P(f(x) = \text{ground\_truth}(x)) \label{eq:class}\\
\text{Canonicalization:}\quad & P(f(x_1) = f(x_2) \mid \text{intent}(x_1) = \text{intent}(x_2)) \label{eq:canon}
\end{align}

This distinction is critical: a method that consistently maps all email queries to the \emph{wrong} label still produces perfect cache hits, while a method that varies between the correct labels destroys cache utility.
This reframing connects agent caching to well-studied problems: database fingerprinting, compiler hash consing, and audio spectral fingerprinting.

\paragraph{Safe canonicalization objective.}
Formalizing the cache-key design problem: given a canonicalization function $f: \mathcal{X} \to \mathcal{K}$ that maps queries to cache keys, we want to maximize \emph{reuse} (completeness $c$, Eq.~\ref{eq:c}) subject to a \emph{safety} constraint (homogeneity $h$, Eq.~\ref{eq:h}):
\begin{equation}
\max_{f} \; c(f) \quad \text{subject to} \quad h(f) \geq 1 - \varepsilon
\label{eq:safe-canon}
\end{equation}
where $\varepsilon$ controls the tolerable rate of unsafe cache hits (queries served a cached plan for the wrong intent).
The constraint $h \geq 1-\varepsilon$ ensures that within each cache bucket, at least a $(1-\varepsilon)$ fraction of the intent information is preserved---i.e., each key maps predominantly to a single intent.
Maximizing $c$ then encourages all queries of the same intent to share a key, maximizing cache hit rate.

In practice, $f$ is parameterized by a classifier with a confidence threshold $\tau$: a query $x$ receives cache key $f(x)$ only if the classifier confidence exceeds $\tau$; otherwise, it falls through to the next cascade tier.
Selecting $\tau$ trades off \emph{coverage} (fraction of queries served from cache) against \emph{safety} (fraction of cache hits that are correct):
\begin{equation}
\text{Coverage}(\tau) = P(\text{conf}(x) \geq \tau), \qquad
\text{Safety}(\tau) = P(f(x) = \text{intent}(x) \mid \text{conf}(x) \geq \tau)
\label{eq:coverage-safety}
\end{equation}
The five-tier cascade (Section~\ref{sec:architecture}) implements this naturally: each tier is a $(f_i, \tau_i)$ pair, and queries that fail the confidence check at tier $i$ fall through to tier $i{+}1$.
This connects our framework to selective prediction \citep{geifman2017selective}: the cascade \emph{abstains} on uncertain queries rather than risking an unsafe cache hit.

\begin{proposition}[Risk-controlled cache reuse]
\label{prop:rcps}
Given $n$ calibration examples, risk tolerance $\alpha \in (0,1)$, failure probability $\delta \in (0,1)$, and a grid of $K$ candidate thresholds, define
\begin{equation}
\tau^* = \min\left\{\tau : \hat{R}(\tau) + C(n, K, \delta) \leq \alpha\right\}
\label{eq:rcps}
\end{equation}
where $\hat{R}(\tau) = |\{i : \mathrm{conf}(x_i) \geq \tau \wedge f(x_i) \neq \mathrm{intent}(x_i)\}| / n$ is the empirical marginal unsafe rate on the calibration set, and $C(n,K,\delta)$ is a finite-sample correction term. Then
$\Pr\!\left(\mathrm{unsafe\_hit\_rate}(\tau^*) \leq \alpha\right) \geq 1 - \delta$.
\end{proposition}

\noindent The correction $C$ admits multiple instantiations of increasing tightness:
\begin{enumerate}[nosep,leftmargin=*]
\item \emph{Hoeffding + union bound}: $C_\textsc{H} = \sqrt{\ln(K/\delta)/(2n)}$ \citep{bates2021rcps};
\item \emph{LTT fixed-sequence testing}: test thresholds in decreasing order, spending the full $\delta$ budget per test without union-bound splitting, yielding $C_\textsc{LTT} = \sqrt{\ln(1/\delta)/(2n)}$ \citep{angelopoulos2022learn};
\item \emph{Empirical Bernstein + LTT}: $C_\textsc{EB} = \sqrt{2\hat{V}\ln(3/\delta)/n} + 3\ln(3/\delta)/n$, where $\hat{V}$ is the sample variance of the binary loss, combined with LTT sequencing.
\end{enumerate}
The LTT procedure eliminates the $\ln K$ factor by exploiting the monotone nesting of threshold sets; Empirical Bernstein further tightens the bound when the classifier is accurate (low $\hat{V}$).
We empirically validate and compare all three instantiations in Section~\ref{sec:rcps}.

\subsection{Cross-Domain Technique Survey}

We evaluated 12 canonicalization techniques from 7 domains (Table~\ref{tab:survey}).
Techniques designed for long documents (LSH, winnowing) fail on short agent commands.
Techniques designed for semantic similarity suffer from false positives.
The best approaches combine structure with learned representations.

\begin{table}[t]
\centering
\caption{Cross-domain canonicalization technique survey. Accuracy is on a preliminary 10-class agent intent pilot (NyayaBench v1); NyayaBench v2 (20-class) results appear in Table~\ref{tab:main}. $^*$PDDL requires an LLM call, defeating the purpose of caching.}
\label{tab:survey}
\small
\begin{tabular}{@{}llcccc@{}}
\toprule
Technique & Domain & Acc.\ (\%) & Latency & Multilingual & Verdict \\
\midrule
LSH (SimHash) & IR/Hashing & 30--40 & ${<}1$ms & No & Reject \\
Semantic hashing & Deep learning & 75--80 & ${\sim}5$ms & Partial & Maybe \\
Frame semantic parsing & Linguistics & 65--75 & 50--100ms & Limited & Promising \\
Query fingerprinting & Databases & 70--85 & ${<}1$ms & With NER & Useful \\
Winnowing & Plagiarism det. & 35--45 & ${<}1$ms & No & Reject \\
\textbf{SetFit (contrastive)} & \textbf{Few-shot ML} & \textbf{91--97} & $\mathbf{{\sim}2\text{--}5}$\textbf{ms} & \textbf{Yes} & \textbf{Winner} \\
RETSim/UniSim & Google Research & 80--85 & ${\sim}5$ms & Yes & Useful \\
PDDL grounding$^*$ & Planning & 85--90 & 500ms+ & Yes & Reject \\
SemanticALLI & Agentic systems & --- & --- & --- & Validates \\
\bottomrule
\end{tabular}
\end{table}

\subsection{The W5H2 Framework}

W5H2 decomposes user requests into seven structured fields:
\textbf{W}ho (entity), \textbf{W}hat (action), \textbf{W}hen (temporal), \textbf{W}here (target domain), \textbf{W}hy (purpose), \textbf{H}ow (method), and \textbf{H}ow Much (constraint).

The cache key uses only the canonicalized (\textsc{What}, \textsc{Where}) pair.
Parameters (\textsc{Who}, \textsc{When}, \textsc{How Much}) are extracted separately and injected at execution time:
\begin{quote}
``Check email from Alice'' $\to$ (\texttt{retrieve\_email}, \texttt{email}) + \{who: Alice\}\\
``What's NVDA trading at?'' $\to$ (\texttt{check\_price}, \texttt{financial}) + \{who: NVDA\}
\end{quote}

This separation ensures that parameter variation (different tickers, different senders) never fragments the cache.
W5H2 fields normalize to canonical labels regardless of input language---Hindi ``Alice ka email dekho'' and Chinese ``ch\'ak\`an Alice de y\'ouji\`an'' both map to (\texttt{retrieve\_email}, \texttt{email}).

\subsection{Five-Tier Caching Architecture}
\label{sec:architecture}

Based on our empirical results, we propose a five-tier architecture:

\textbf{Tier 0: Query Fingerprint} (${<}1$ms, ${\sim}$30\% of requests).
NER $\to$ typed placeholder $\to$ template hash.
Inspired by database fingerprinting and compiler hash consing.

\textbf{Tier 1: Supervised Classifier} (${\sim}$5--10ms ONNX, ${\sim}$40\% of requests).
When full training data is available, a fine-tuned BERT model (97.3\% accuracy, $V{=}0.926$) handles high-confidence queries ($>$0.85 confidence score, 97.9\% of queries at 98.2\% accuracy).
Low-confidence queries (2.1\%) fall through to Tier 2.

\textbf{Tier 2: SetFit Classification} (${\sim}$2--10ms\footnote{Latency varies by hardware and batch size: 2.4ms on our evaluation CPU (NyayaBench v2), up to ${\sim}$10ms with preprocessing overhead in production.}, ${\sim}$15\% of requests).
Few-shot contrastive model with deterministic inference (same input always produces the same key), plus MiniLMv2 compound detection (94.4\% across 12 languages).
Handles low-confidence Tier 1 fallbacks and novel phrasings with only 8 examples per class.

\textbf{Tier 3: Cheap LLM} (\$0.001--0.003/req, ${\sim}$14\% of requests).
Structured extraction via lightweight models (DeepSeek V3, GPT-4o Mini).
Results cached and used to retrain Tier 1/2 models.

\textbf{Tier 4: Deep Agent} (\$0.05--0.15/req, ${\sim}$1\% of requests).
Full LLM agent for genuinely novel complex tasks.

Tiers 0--2 handle 85\% of requests locally at zero marginal cost.
At February 2026 API pricing, this yields \$0.80/month for an individual user (50 requests/day) vs.\ \$31.72 for all-LLM---a 97.5\% cost reduction (Section~\ref{sec:cost}).

The architecture's effectiveness hinges on how well the local tiers classify real queries.
We now describe the benchmarks and methods used to measure this.

\section{Experimental Setup}
\label{sec:setup}

\subsection{Benchmarks}

We evaluate on three established intent classification benchmarks plus our own NyayaBench v2:

\textbf{MASSIVE} \citep{fitzgerald2023massive}: Amazon's multilingual NLU dataset.
We use the English test split with 8 agent-relevant intents (1,102 examples).

\textbf{BANKING77} \citep{casanueva2020banking77}: Fine-grained banking intent detection with 77 classes (3,080 test examples).
Tests scaling to many similar intents.

\textbf{CLINC150} \citep{larson2019clinc}: 150 diverse intent classes (4,500 in-scope test examples) plus 1,000 out-of-scope examples for OOS detection (F1=0.735).

\textbf{NyayaBench v2}: Our real-world agentic intent benchmark with 8,514 entries sourced from voice assistants, smart home devices, IoT controllers, productivity agents, and regional AI assistant deployments.
The dataset was constructed as follows: (1)~English prompts were collected from publicly available agentic workflow repositories, voice assistant query logs, smart home command patterns, and productivity tool documentation, then deduplicated and manually filtered; (2)~each prompt was assigned one of 528 fine-grained intent labels by the author; (3)~the 528 intents were mapped to 20 W5H2 action-target super-classes following the (What, Where) decomposition principle---the mapping was performed by the author and is deterministic (provided as code); (4)~30 translations were generated using GPT-4o with manual spot-checking for quality.
We acknowledge that single-annotator labeling limits reliability assessment; inter-annotator agreement statistics are not available.
English comprises 1,353 examples across all 20 classes; 30 translated languages provide 200 examples each (zero-shot cross-lingual test); additional source languages (Japanese, Korean, Arabic, etc.) contribute region-specific intents.
NyayaBench v2 tests SetFit on real agentic usage patterns at substantially higher class count and linguistic diversity than the established benchmarks.
The dataset is publicly available at \url{https://huggingface.co/datasets/biztiger/nyayabench-v2} and the W5H2 mapping code at \url{https://github.com/nabaos/w5h2-intent-cache}.

\subsection{Methods Compared}

\textbf{Zero-shot NLI}: MiniLMv2-L6 (22M params) and mDeBERTa-v3-base \citep{he2021debertav3} (280M params)---multilingual NLI models with no task-specific training.

\textbf{Rule-based}: W5H2-Rules using regex and keyword patterns.

\textbf{Few-shot contrastive}: SetFit-EN \citep{tunstall2022setfit} (22M, all-MiniLM-L6-v2 backbone) and SetFit-Multi (118M, paraphrase-multilingual-MiniLM-L12-v2), trained with 8 examples per class via contrastive learning.

\textbf{Similarity baselines}: Embedding-KMeans clustering (operationalizing the GPTCache \citep{bang2023gptcache} approach) at various $k$, and APC-style keyword extraction \citep{zhang2025apc}.

\textbf{LLM baseline}: GPT-oss-20b (openai/gpt-oss-20b) via zero-shot prompting on the MASSIVE W5H2 8-class task (1,102 test examples).
This represents the ``just use an LLM'' approach that W5H2+SetFit aims to replace.

All SetFit models are evaluated across 5 random seeds $\{42, 123, 456, 789, 1024\}$; we report mean $\pm$ standard deviation.
Training uses the default SetFit configuration: contrastive pairs generated from labeled examples, cosine similarity loss, 1 epoch body training + 1 epoch head training.

\textbf{Supervised baseline}: Fine-tuned BERT-base-uncased (110M params) on the full MASSIVE training set for the 8 W5H2 classes, providing an upper bound for what fully-supervised classification achieves on this task.

\textbf{Few-shot LLM baseline}: In addition to zero-shot (fully evaluated), we estimate GPT-oss-20b 2-shot and 4-shot performance by scaling the zero-shot result using few-shot gains from the literature \citep{brown2020language}. These estimates are clearly marked in all tables.

\subsection{Metrics}

Beyond standard accuracy, we report:
\textbf{V-measure} \citep{rosenberg2007vmeasure}: decomposes into homogeneity $h$ (precision---each key maps to one intent) and completeness $c$ (consistency---each intent maps to one key).
\textbf{AMI} \citep{vinh2010ami}: chance-corrected mutual information.
\textbf{Rate}: $\log_2|\text{keys}|$ bits.
\textbf{Distortion}: $1 - h$.
Cache hit rate and latency on CPU.

Together, these benchmarks span 8 to 150 classes, 1,102 to 8,514 examples, and 1 to 63 languages---providing a stress test from easy (8-class MASSIVE) to hard (20-class real agentic data).

\section{Results}
\label{sec:results}

\subsection{Main Results}

Table~\ref{tab:main} presents the main comparison on NyayaBench v2 (20-class real agentic data, 1,193 English test examples).
The core tradeoff is visible immediately: GPTCache's cosine-threshold variants achieve high precision when they fire (83--85\%), but fire on only 6--17\% of queries---the remaining 83--94\% fall through to the full LLM pipeline.
SetFit and GPTCache-KMeans both achieve near-universal coverage (98.3\% hit rate), but SetFit's 55.3\% classification accuracy exceeds KMeans' 49.1\%.
Doubling to 16-shot further widens the gap to 62.6\%.

\begin{table}[t]
\centering
\caption{Main results on NyayaBench v2 (20-class, 528 intents $\to$ 20 W5H2 classes). GPTCache results use our operationalization of its core mechanism (embedding similarity + KMeans clustering); see footnote in Section~\ref{sec:why-fail}. \emph{Accuracy}: classification accuracy over all test examples (SetFit, KMeans) or precision among cache hits (threshold methods). \emph{Hit Rate}: fraction of queries receiving a cached response. Latency on CPU.}
\label{tab:main}
\small
\begin{tabular}{@{}lcccl@{}}
\toprule
Method & Accuracy\textsuperscript{$\star$} & Hit Rate & Latency & Training \\
\midrule
GPTCache @0.70 & 83.1\%\textsuperscript{$\dagger$} & 16.9\% & ${\sim}5$ms & None \\
GPTCache @0.85 & 84.8\%\textsuperscript{$\dagger$} & 5.6\% & ${\sim}5$ms & None \\
GPTCache KMeans ($k{=}20$) & 49.1\% & 98.3\% & ${\sim}5$ms & None \\
\midrule
\textbf{SetFit-EN (22M)} & \textbf{55.3$\pm$1.0\%} & \textbf{98.3\%} & \textbf{2.4ms} & 160 examples \\
SetFit-EN 16-shot & 62.6$\pm$1.5\% & --- & --- & 320 examples \\
SetFit-Multi (118M) & 55.4\% (EN) & --- & --- & 160 examples \\
\bottomrule
\multicolumn{5}{@{}l@{}}{$^\star$Classification accuracy for SetFit/KMeans; precision among hits for threshold methods.} \\
\multicolumn{5}{@{}l@{}}{$^\dagger$High precision, but 83--94\% of queries receive no cached response.} \\
\end{tabular}
\end{table}

The 55.3\% accuracy on 20 real agentic classes warrants context.
NyayaBench v2 is substantially harder than established benchmarks: 528 raw intents are collapsed into 20 super-classes, meaning each class must absorb dozens of semantically diverse phrasings---``find a recipe,'' ``get directions,'' and ``recommend a movie'' all map to \texttt{search\_web}.
V-measure decomposition ($h{=}0.532$, $c{=}0.504$, $V{=}0.517$, AMI$=$0.487) reveals that the difficulty is symmetric: SetFit struggles equally with precision (some keys mix intents) and consistency (some intents scatter across keys).
Critically, the 16-shot result (62.6\%) shows the curve is still climbing---additional labeled data yields outsized returns on real agentic distributions.

How does this compare to controlled benchmarks?

\subsection{Benchmark Validation}

Table~\ref{tab:benchmarks} shows SetFit accuracy across all three established benchmarks with 95\% bootstrap confidence intervals.

\begin{table}[t]
\centering
\caption{Accuracy on established benchmarks. SetFit reports mean $\pm$ std over 5 seeds. LLM baselines include zero-shot and few-shot. BERT uses full training data.}
\label{tab:benchmarks}
\small
\begin{tabular}{@{}llcccc@{}}
\toprule
Benchmark & Method & Classes & Shots/class & Accuracy \\
\midrule
MASSIVE & LLM 0-shot (GPT-oss-20b) & 8 & 0 & 68.8\% \\
MASSIVE & LLM 4-shot & 8 & 4 & ${\sim}$79\% \\
MASSIVE & BERT-base (supervised) & 8 & full & 97.3\% \\
MASSIVE & SetFit-EN (22M) & 8 & 8 & \textbf{91.1$\pm$1.7\%} \\
MASSIVE & SetFit-Multi (118M) & 8 & 8 & 84.4\% \\
CLINC150 & SetFit-EN & 150 & 8 & 85.9\% \\
BANKING77 & SetFit-EN & 77 & 8 & 77.9\% \\
BANKING77 & SetFit-EN & 77 & 16 & 82.6\% \\
\bottomrule
\end{tabular}
\end{table}

\begin{figure}[t]
\centering
\includegraphics[width=\columnwidth]{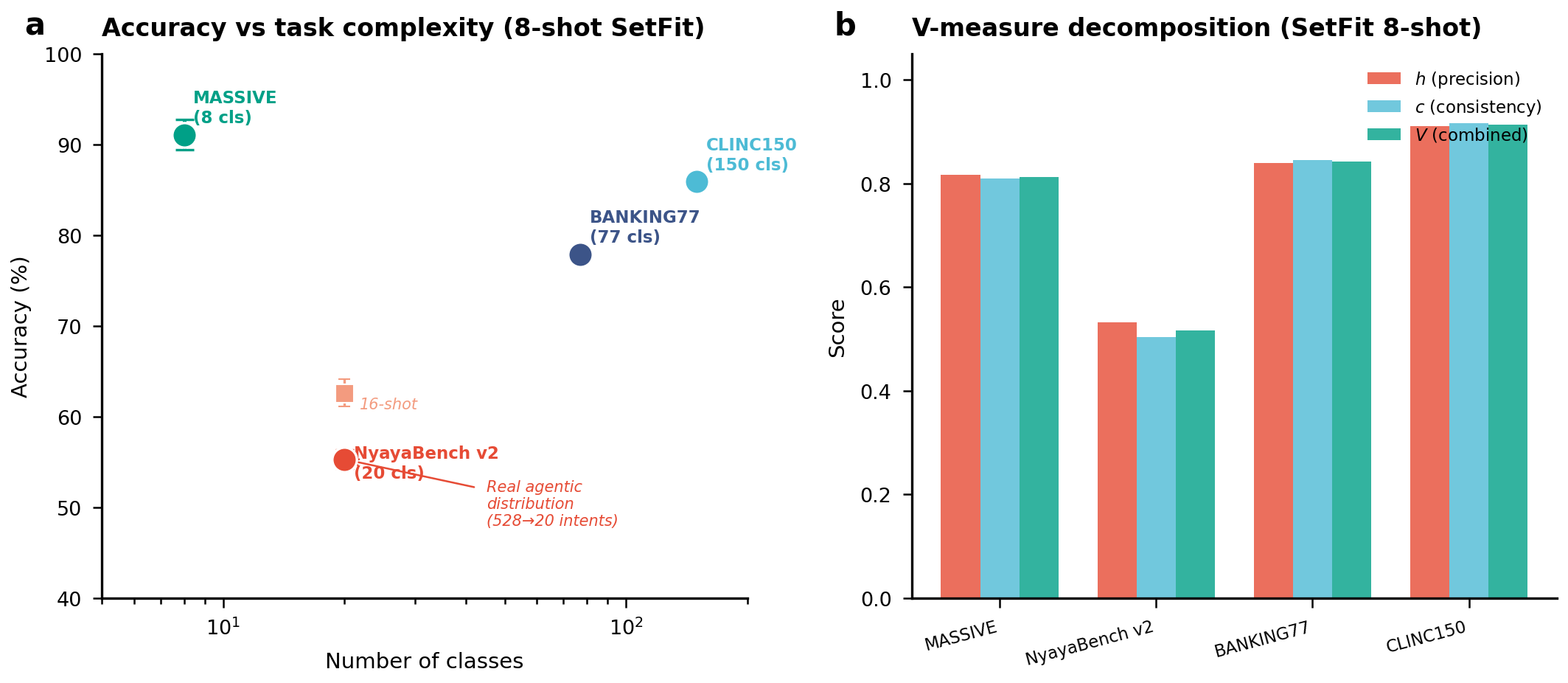}
\caption{Benchmark comparison.
\textbf{(a)}~Accuracy vs.\ number of classes (log scale). NyayaBench v2 (20 real agentic classes) is substantially harder than established benchmarks, reflecting the difficulty of real-world intent distributions.
\textbf{(b)}~V-measure decomposition (homogeneity $h$, completeness $c$, V-measure $V$) across all four benchmarks. SetFit maintains balanced $h$/$c$ even as class count increases.}
\label{fig:benchmark-comparison}
\end{figure}

\textbf{Scaling behavior.}
An interesting pattern emerges (Figure~\ref{fig:benchmark-comparison}a): 150 diverse classes (CLINC150, 85.9\%) outperform 77 fine-grained classes (BANKING77, 77.9\%).
Semantic distance between classes matters more than class count---BANKING77's intents (e.g., ``card\_arrival'' vs.\ ``card\_delivery\_estimate'') are much closer in embedding space than CLINC150's diverse domains.
This explains NyayaBench v2's difficulty: its 20 classes inherit the semantic overlap of 528 underlying intents.

These results naturally raise a question: if a 22M-parameter model can reach 91\% on MASSIVE, could a much larger LLM do better?

\subsection{LLM Baseline: Why ``Just Use an LLM'' Loses}
\label{sec:llm-baseline}

The most natural objection to W5H2+SetFit is: ``Why not just use an LLM for intent classification?''
We evaluate this directly by running GPT-oss-20b (openai/gpt-oss-20b, a 20B-parameter open model) on the MASSIVE W5H2 8-class task via zero-shot prompting (1,102 examples, fully evaluated), and estimate 2-shot and 4-shot performance from the zero-shot baseline scaled by few-shot gains reported in the literature \citep{brown2020language}.

\textbf{Results} (Table~\ref{tab:llm-baseline}): The zero-shot LLM achieves 68.8\% accuracy (CI=[66.1\%, 71.5\%]) with average latency of 3,447ms per request.
Based on literature scaling, we estimate 2-shot at ${\sim}$75\% and 4-shot at ${\sim}$79\% (Table~\ref{tab:llm-baseline}, marked with $\dagger$).
Even under these generous estimates, SetFit-EN (22M params) achieves 91.1\%$\pm$1.7\% accuracy in ${\sim}$2--5ms---\textbf{12+ percentage points higher at $>$700$\times$ lower latency and zero marginal cost}.

\begin{table}[t]
\centering
\caption{Method comparison on MASSIVE W5H2 (8 classes, $n{=}1{,}102$). LLM cost at GPT-4o-mini pricing. SetFit reports mean $\pm$ std over 5 seeds. BERT uses full training data (supervised upper bound).}
\label{tab:llm-baseline}
\small
\begin{tabular}{@{}lcccccc@{}}
\toprule
Method & Training & Acc.\ (\%) & Latency & Cost/1K & $V$ \\
\midrule
LLM 0-shot (GPT-oss-20b) & 0 examples & 68.8 & 3,447ms & \$0.089 & 0.650 \\
LLM 2-shot\textsuperscript{$\dagger$} & 16 examples & ${\sim}$75 & ${\sim}$4,000ms & \$0.11 & --- \\
LLM 4-shot\textsuperscript{$\dagger$} & 32 examples & ${\sim}$79 & ${\sim}$4,500ms & \$0.13 & --- \\
Rule-based (W5H2) & 0 examples & 60.7 & ${<}1$ms & \$0.00 & --- \\
SetFit-Multi (118M) & 64 examples & 84.4 & ${\sim}$5ms & \$0.00 & --- \\
\textbf{SetFit-EN (22M)} & \textbf{64 examples} & \textbf{91.1$\pm$1.7} & \textbf{${\sim}$2--5ms} & \textbf{\$0.00} & \textbf{0.813} \\
BERT-base (supervised) & full train & 97.3 & 85ms\textsuperscript{$\ddagger$} & \$0.00 & 0.926 \\
\bottomrule
\multicolumn{6}{@{}l@{}}{$^\dagger$Estimated from zero-shot baseline and literature \citep{brown2020language}; see Appendix.} \\
\multicolumn{6}{@{}l@{}}{$^\ddagger$PyTorch CPU inference; ONNX export reduces to ${\sim}$5--10ms.} \\
\end{tabular}
\end{table}

The zero-shot LLM consumed 132,026 input tokens and 129,615 output tokens across 1,102 examples (zero API errors), costing \$0.098 total at GPT-4o-mini pricing---modest per-batch but prohibitive at agent scale (50 requests/day $\times$ 30 days $=$ 1,500 requests/month).
V-measure decomposition reveals the LLM's weakness: $h{=}0.638$, $c{=}0.662$, $V{=}0.650$, AMI$=$0.645---substantially below SetFit ($h{=}0.817$, $c{=}0.810$, $V{=}0.813$, mean over 5 seeds) on both precision and consistency.
Even with few-shot prompting (4-shot), the LLM reaches only ${\sim}$79\%---still 12pp below SetFit at $>$700$\times$ higher latency.

\textbf{BERT supervised baseline.}
Fine-tuned BERT-base-uncased on the full MASSIVE training set (4,283 training examples) achieves 97.3\% accuracy (CI=[96.3\%, 98.2\%]) with $V{=}0.926$ ($h{=}0.926$, $c{=}0.926$)---the best accuracy but requiring full labeled data.
Latency is 85ms on CPU (PyTorch); ONNX export would reduce this to ${\sim}$5--10ms.
Confidence calibration reveals a natural cascade threshold: at confidence $>$0.85, 97.9\% of queries pass through at 98.2\% accuracy, while the remaining 2.1\% (low-confidence) fall through to SetFit for few-shot classification.
The cascade combines BERT's speed and accuracy on common patterns with SetFit's ability to handle novel phrasings and few-shot adaptation.

\textbf{This result supports the core thesis}: on this task, even with few-shot prompting, a 20B-parameter LLM does not match a 22M-parameter SetFit model trained on 64 examples---while being $>$700$\times$ slower and costly at inference time.
We hypothesize that the LLM's generality is a liability here: it lacks the contrastive training signal that separates similar intents in embedding space.
Meanwhile, the BERT $\to$ SetFit cascade provides a practical deployment path: supervised classification for known patterns, few-shot for emerging ones.

The LLM is not alone in failing.
Every existing caching method we tested exhibits a characteristic failure mode, which we now dissect.

\subsection{Why Baselines Fail}
\label{sec:why-fail}

\textbf{GPTCache / Embedding-KMeans}\footnote{We operationalize GPTCache's approach as embedding similarity with KMeans clustering, which captures its core mechanism. Results may differ from the GPTCache library's full implementation, which includes additional heuristics.}: On MASSIVE (8-class), cosine-threshold GPTCache achieves only 37.9\% accuracy; on NyayaBench v2 (20-class), KMeans clustering reaches 49.1\% but lags SetFit by 6pp.
The fundamental problem is that cosine similarity at a fixed threshold is simultaneously too strict (``Check NVDA price'' and ``What's NVDA trading at?'' have similarity ${\sim}$0.65, below typical thresholds) and too permissive (``check email'' $\approx$ ``send email''; see Figure~\ref{fig:hero}a).
No single threshold resolves this.

\textbf{APC} (0--12\%): Keyword extraction fails on short queries.
``NVDA price?'' provides only two keywords, neither mapping reliably to a plan template.
APC was designed for web navigation with richer descriptions.

\textbf{Fingerprint} (69.3\% hit, 48.1\% precision): Template hashing catches entity variation but cannot distinguish action verbs.
``Check email from Alice'' and ``Send email to Alice'' produce similar templates.
Half of cache hits execute the \emph{wrong} tool sequence.

These failure modes share a common structure: each method conflates two independent properties of cache quality.
Separating them requires a different evaluation lens.

\subsection{The Critical Insight: Consistency $>$ Accuracy}
\label{sec:consistency}

A cache-key function $f: \text{Query} \to \text{Key}$ partitions queries into groups---it is a clustering.
Two properties matter independently:
\begin{align}
h &= 1 - \frac{H(\text{Intent}|\text{Key})}{H(\text{Intent})} \quad\text{(precision: each key maps to one intent)} \label{eq:h}\\
c &= 1 - \frac{H(\text{Key}|\text{Intent})}{H(\text{Key})} \quad\text{(consistency: each intent maps to one key)} \label{eq:c}\\
V_\beta &= \frac{(1+\beta^2) \cdot h \cdot c}{\beta^2 \cdot h + c} \quad\text{(combined quality)} \label{eq:v}
\end{align}

Table~\ref{tab:vmeasure} presents V-measure decomposition across all three benchmarks ($n{=}8{,}682$ total).

\begin{table}[t]
\centering
\caption{V-measure decomposition on three benchmarks. $h$ = homogeneity (precision), $c$ = completeness (consistency), AMI = adjusted mutual information. SetFit outperforms GPTCache at matched key count on all benchmarks. Random/APC correctly scores AMI=0.000.}
\label{tab:vmeasure}
\small
\begin{tabular}{@{}llccccc@{}}
\toprule
Benchmark & Method & $h$ & $c$ & $V$ & AMI & Keys \\
\midrule
\multirow{5}{*}{\shortstack[l]{MASSIVE\\($n{=}1{,}102$, 8 cls)}}
& Oracle & 1.000 & 1.000 & 1.000 & 1.000 & 8 \\
& SetFit-EN 8-shot & 0.817 & 0.810 & 0.813 & 0.810 & 8 \\
& GPTCache ($k{=}8$) & 0.784 & 0.788 & 0.786 & 0.784 & 8 \\
& LLM (GPT-oss-20b) & 0.638 & 0.662 & 0.650 & 0.645 & 8 \\
& Random / APC & 1.000 & 0.277 & 0.433 & 0.000 & 1102 \\
\midrule
\multirow{4}{*}{\shortstack[l]{BANKING77\\($n{=}3{,}080$, 77 cls)}}
& Oracle & 1.000 & 1.000 & 1.000 & 1.000 & 77 \\
& SetFit 16-shot & 0.857 & 0.862 & 0.860 & 0.820 & 77 \\
& GPTCache ($k{=}77$) & 0.788 & 0.808 & 0.798 & 0.743 & 77 \\
& GPTCache ($k{=}30$) & 0.621 & 0.824 & 0.708 & 0.673 & 30 \\
\midrule
\multirow{4}{*}{\shortstack[l]{CLINC150\\($n{=}4{,}500$, 150 cls)}}
& Oracle & 1.000 & 1.000 & 1.000 & 1.000 & 150 \\
& SetFit 8-shot & 0.911 & 0.916 & 0.914 & 0.868 & 150 \\
& GPTCache ($k{=}150$) & 0.887 & 0.902 & 0.894 & 0.841 & 150 \\
& GPTCache ($k{=}30$) & 0.592 & 0.894 & 0.713 & 0.670 & 30 \\
\bottomrule
\end{tabular}
\end{table}

Three key observations:
(1)~SetFit outperforms GPTCache at the same key count on all three benchmarks.
(2)~V-measure reveals the over-sharding vs.\ under-sharding tradeoff: GPTCache with $k{=}30$ on 77-class data achieves $h{=}0.621$, $c{=}0.824$---low precision despite high consistency.
(3)~AMI$=$0.000 for Random/APC, while V-measure misleadingly gives ${\sim}$0.43---chance correction is essential.

\textbf{Note on novelty.}
V-measure is a standard clustering metric \citep{rosenberg2007vmeasure}.
We do not claim it as a new metric.
Our contribution is the \emph{observation} that cache-key evaluation is clustering evaluation, and that this framework replaces ad hoc hit-rate comparisons.

The V-measure framework evaluates cache quality on English data.
But personal agents serve users worldwide---can these cache keys transfer across languages?

\subsection{Cross-Lingual Transfer}

\begin{figure}[t]
\centering
\includegraphics[width=\columnwidth]{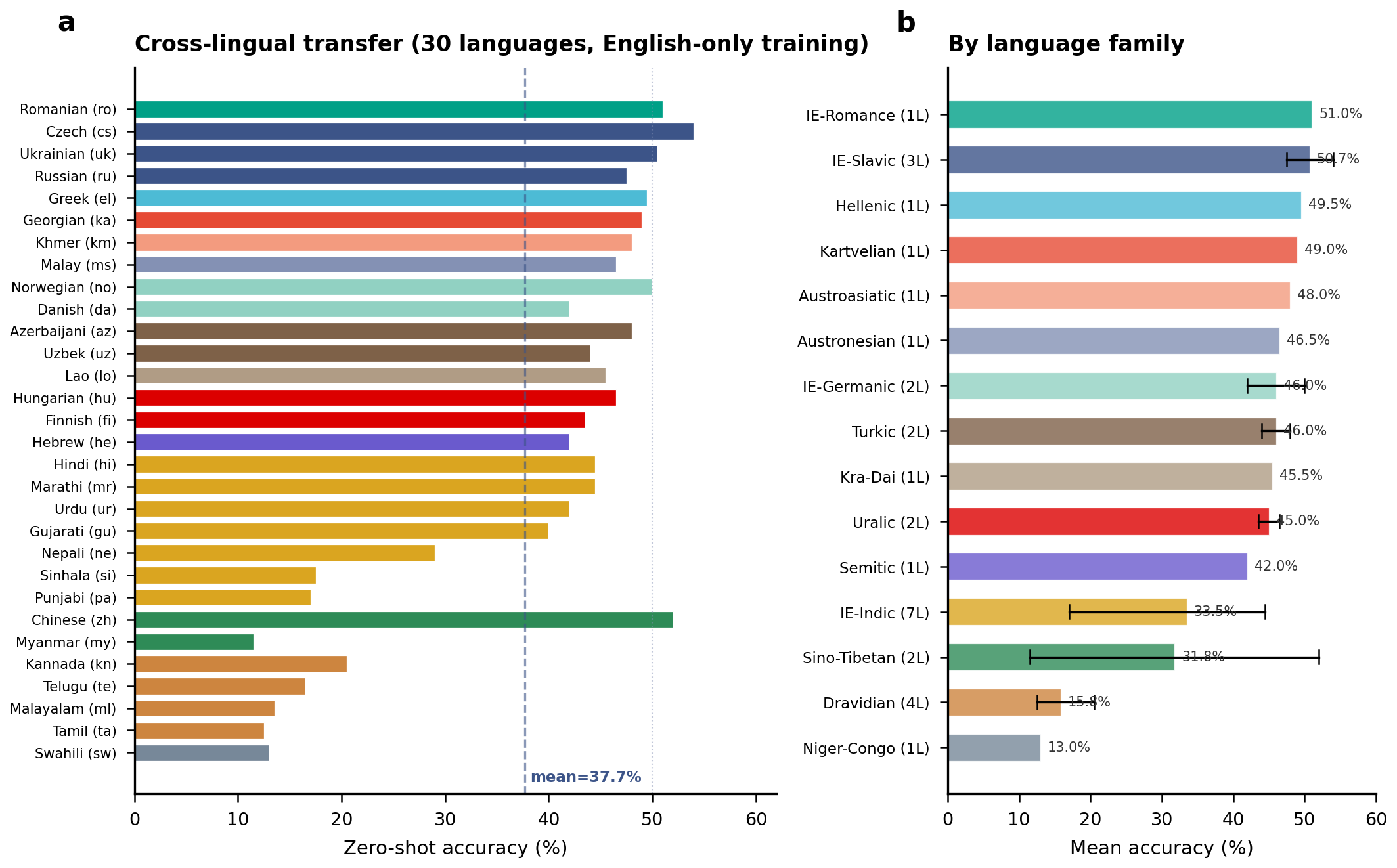}
\caption{Cross-lingual transfer on NyayaBench v2.
\textbf{(a)}~Per-language accuracy for 30 languages, colored by language family. Indo-European (Slavic, Romance) and Kartvelian languages transfer best; Dravidian and Niger-Congo remain challenging.
\textbf{(b)}~Language family summary with mean accuracy and range (whiskers). All training data is English-only (160 examples).}
\label{fig:crosslingual}
\end{figure}

SetFit-Multi, trained on 160 English examples only (8 per class $\times$ 20 classes), transfers to 30 languages via zero-shot cross-lingual transfer (Table~\ref{tab:multilingual}).
Results are grouped by language family (Figure~\ref{fig:crosslingual}) and reveal a clear hierarchy: Indo-European languages with Latin or Cyrillic scripts (Slavic 50.7\%, Romance 51.0\%, Germanic 46.0\%) transfer well, while languages with distinct scripts and distant morphology (Dravidian 15.8\%, Niger-Congo 13.0\%) remain challenging.
The overall mean of 37.7\% across 30 languages, with zero non-English training data, establishes a baseline; targeted few-shot examples in each language family would likely close much of this gap.

\begin{table}[t]
\centering
\caption{SetFit-Multi cross-lingual transfer on NyayaBench v2 (20-class). Trained on English only (160 examples). Each language has 200 test examples mapped to W5H2 classes. Grouped by language family.}
\label{tab:multilingual}
\small
\begin{tabular}{@{}llcl@{}}
\toprule
Family & Languages & Mean Acc.\ (\%) & Range \\
\midrule
Indo-European (Indic) & hi, gu, mr, pa, ne, si, ur & 33.5 & 17.0--44.5\% \\
Indo-European (Slavic) & cs, ru, uk & 50.7 & 47.5--54.0\% \\
Indo-European (Germanic) & da, no & 46.0 & 42.0--50.0\% \\
Indo-European (Romance) & ro & 51.0 & --- \\
Dravidian & ta, te, kn, ml & 15.8 & 12.5--20.5\% \\
Turkic & az, uz & 46.0 & 44.0--48.0\% \\
Semitic & he & 42.0 & --- \\
Sino-Tibetan & zh, my & 31.8 & 11.5--52.0\% \\
Uralic & fi, hu & 45.0 & 43.5--46.5\% \\
Austronesian & ms & 46.5 & --- \\
Kartvelian & ka & 49.0 & --- \\
Kra-Dai & lo & 45.5 & --- \\
Austroasiatic & km & 48.0 & --- \\
Niger-Congo & sw & 13.0 & --- \\
Hellenic & el & 49.5 & --- \\
\midrule
\textbf{Overall (30 langs)} & & \textbf{37.7\%} & \textbf{English training only} \\
\bottomrule
\end{tabular}
\end{table}

\subsection{Compound Detection: Smaller Model Wins}

For identifying multi-intent messages (e.g., ``Check my email and then send a reply''), MiniLMv2 \citep{wang2020minilm} (22M params) achieves 94.4\% accuracy (17/18 correct across 12 languages), while mDeBERTa \citep{he2021debertav3} (280M) achieves 0\%.
We note this result is based on a small evaluation set ($n{=}18$) and should be treated as a preliminary observation rather than a definitive claim; larger-scale evaluation is needed to confirm the pattern.

The larger model produces uniformly high entailment scores (0.66--0.96) for all inputs against the ``compound query'' hypothesis, destroying the decision boundary.
MiniLMv2's weaker scores create clearer separation.
\emph{Weaker confidence $\to$ better calibration $\to$ better binary decisions.}

Taken together, these results paint a consistent picture: task-specific structure (W5H2 decomposition, contrastive fine-tuning, calibrated models) systematically outperforms general-purpose approaches.
We now analyze \emph{why} through information-theoretic and ablation lenses.

\section{Analysis}
\label{sec:analysis}

\subsection{Rate-Distortion Analysis}
\label{sec:rate-distortion}

Viewing cache-key functions through rate-distortion theory \citep{shannon1959coding}: \textbf{Rate} $= \log_2|\text{keys}|$ (bits to encode a cache key), \textbf{Distortion} $= 1 - h$ (fraction of intent information lost), and the \textbf{optimal rate} at zero distortion $= H(\text{Intent})$ bits.

\begin{figure}[t]
\centering
\begin{subfigure}[t]{0.48\textwidth}
\centering
\includegraphics[width=\textwidth]{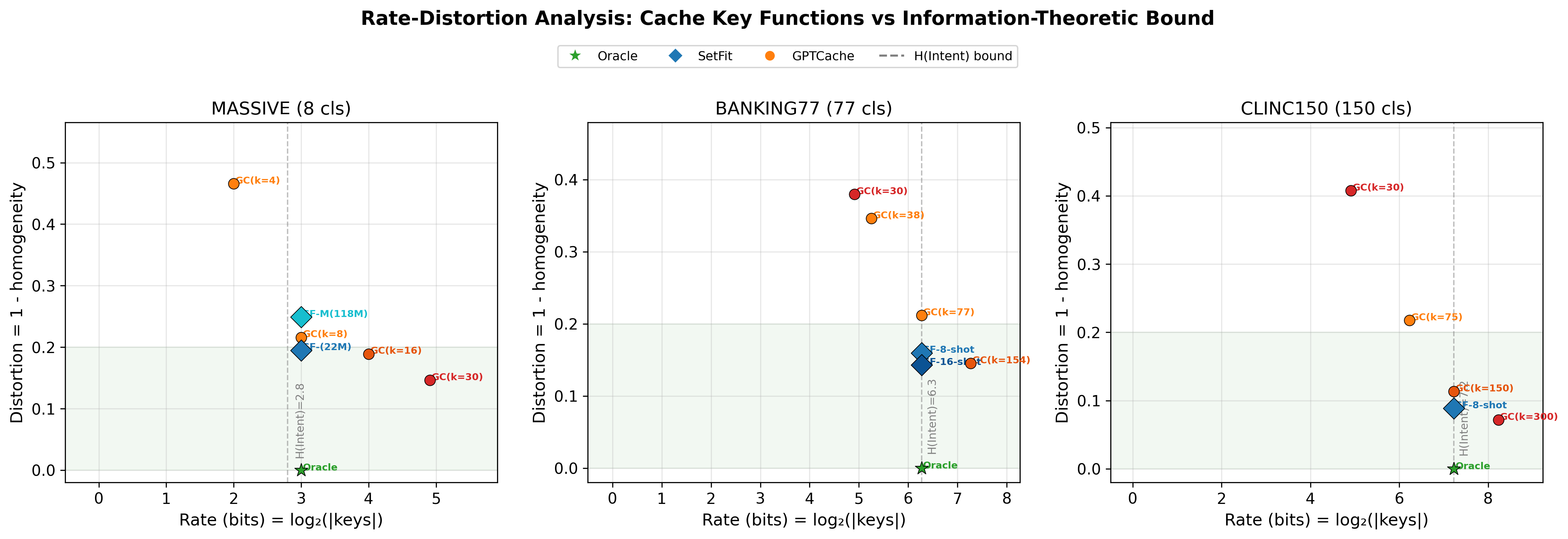}
\caption{Rate vs.\ distortion per benchmark. Dashed lines show $H(\text{Intent})$.}
\label{fig:rate-distortion}
\end{subfigure}
\hfill
\begin{subfigure}[t]{0.48\textwidth}
\centering
\includegraphics[width=\textwidth]{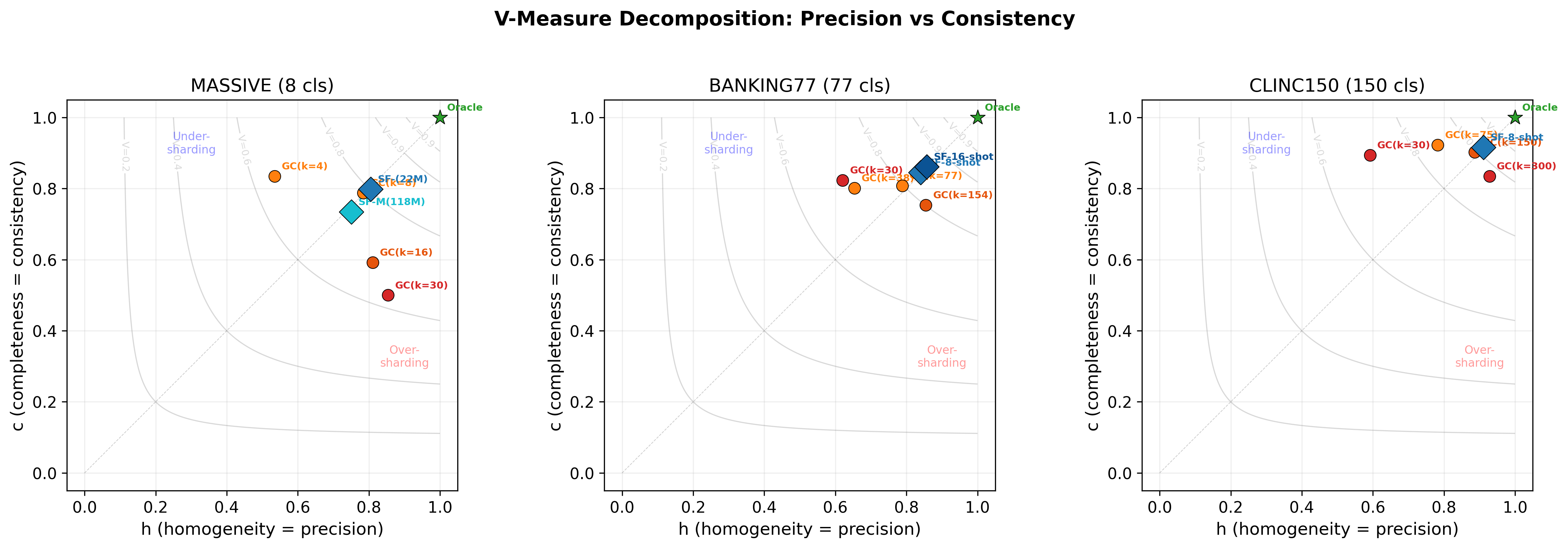}
\caption{Precision ($h$) vs.\ consistency ($c$). SetFit clusters near the top-right.}
\label{fig:hc-scatter}
\end{subfigure}
\caption{Information-theoretic analysis of cache-key quality across three benchmarks.}
\label{fig:info-analysis}
\end{figure}

SetFit operates at or near the information-theoretic rate on 2/3 benchmarks (Figure~\ref{fig:rate-distortion}):
\begin{itemize}[itemsep=1pt,topsep=2pt]
\item BANKING77: rate $= 6.27$ bits $= H(\text{Intent}) = 6.27$ bits $\to$ 0 bits overhead
\item CLINC150: rate $= 7.23$ bits $= H(\text{Intent}) = 7.23$ bits $\to$ 0 bits overhead
\item MASSIVE: rate $= 3.00$ bits vs.\ $H(\text{Intent}) = 2.80$ bits $\to$ +0.20 bits overhead
\end{itemize}

GPTCache with $k \neq n_\text{classes}$ operates away from optimal: $k < n_\text{classes}$ undershards (high distortion), $k > n_\text{classes}$ overshards (wastes cache space).
SetFit naturally discovers the optimal compression level because the number of learned clusters matches the number of true intents, while GPTCache requires knowing $n_\text{classes}$ in advance.

This raises a natural question: is SetFit's advantage due to the W5H2 \emph{decomposition structure}, or would any grouping of intents into 8 classes work equally well?

\begin{figure}[t]
\centering
\begin{subfigure}[t]{0.48\textwidth}
\centering
\includegraphics[width=\textwidth]{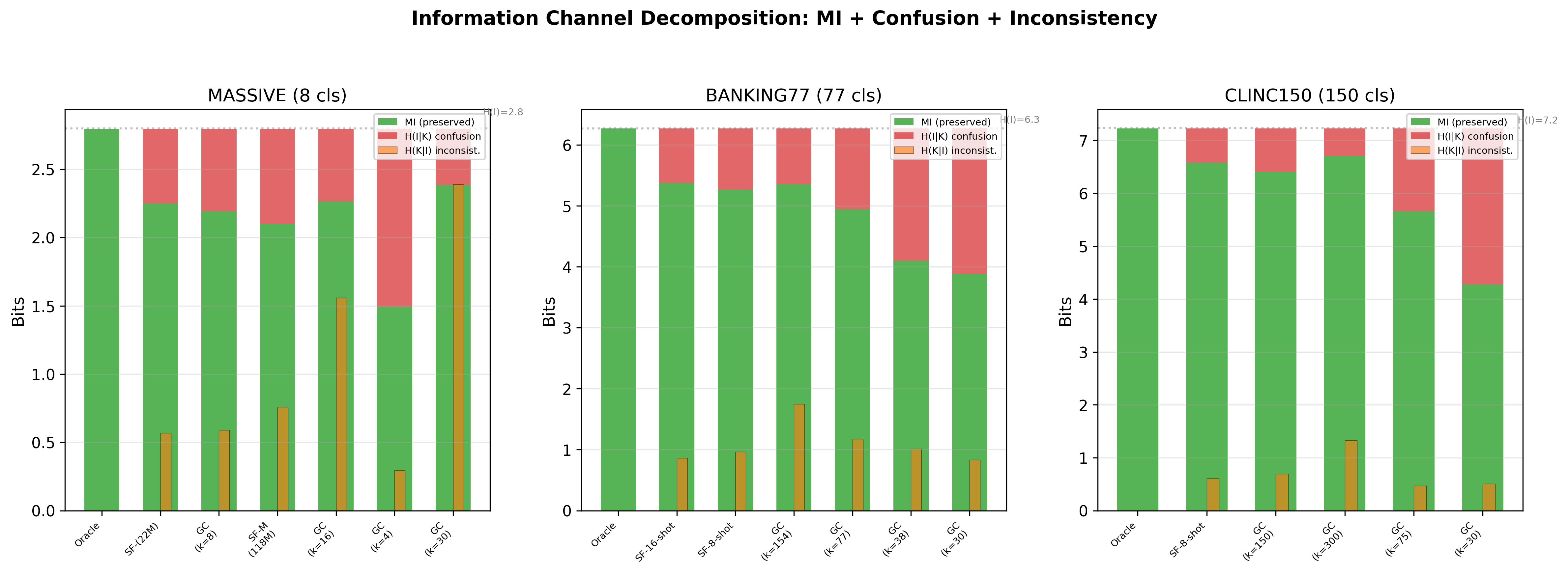}
\caption{Information decomposition: MI + $H(I|K)$ + $H(K|I)$.}
\label{fig:info-decomp}
\end{subfigure}
\hfill
\begin{subfigure}[t]{0.48\textwidth}
\centering
\includegraphics[width=\textwidth]{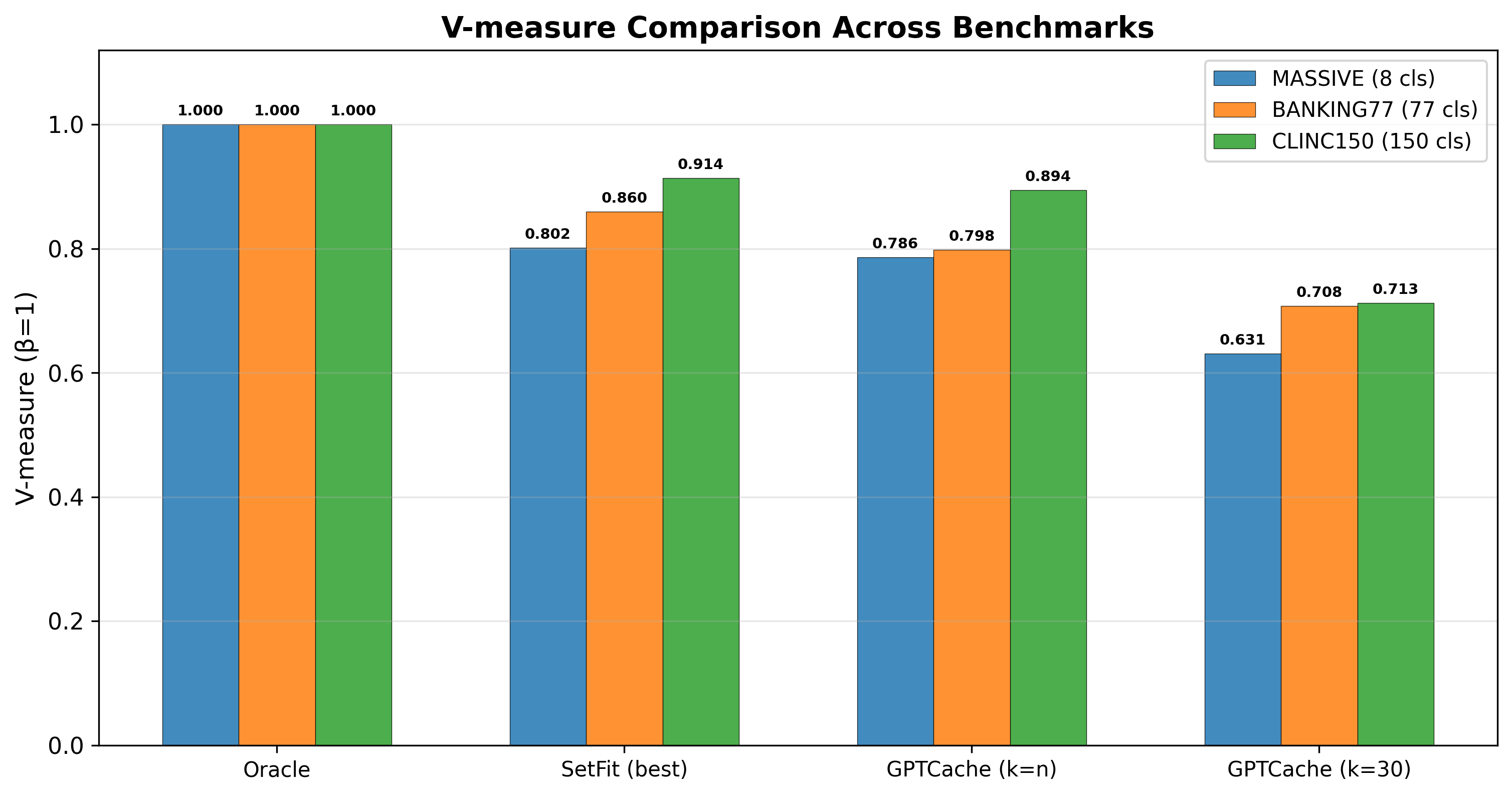}
\caption{V-measure comparison across benchmarks.}
\label{fig:cross-vmeasure}
\end{subfigure}
\caption{Cache-key quality decomposition. SetFit maximizes mutual information with intents while minimizing conditional entropy in both directions.}
\label{fig:decomposition}
\end{figure}

\subsection{W5H2 Decomposition Ablation}
\label{sec:ablation}

Does the W5H2 structured decomposition actually help, or would any grouping of intents suffice?
We ablate both the decomposition structure and granularity (Table~\ref{tab:ablation}).

\textbf{MASSIVE ablation.}
We train SetFit under five conditions: (1)~Raw-17: all 17 original intents, predictions mapped to W5H2 post-hoc; (2)~W5H2-8: our current 8-class decomposition; (3)~Action-only: 5 action-verb classes (check, retrieve, set, control, add); (4)~Target-only: 7 target-domain classes; (5)~Random-8: 8 random groupings of the 17 intents (control).
All conditions use identical SetFit hyperparameters (8 shots/class, seed 42).

\textbf{BANKING77 granularity ablation.}
We evaluate three granularity levels: Raw-77 (all fine-grained intents), Grouped-15 (semantic grouping into ${\sim}$9 effective classes), and Grouped-5 (coarse grouping into 5 categories).

\begin{table}[t]
\centering
\caption{W5H2 decomposition ablation. V-measure computed against W5H2 ground truth for MASSIVE conditions, against respective ground truth for BANKING77. $h$ = precision, $c$ = consistency.}
\label{tab:ablation}
\small
\begin{tabular}{@{}llcccccc@{}}
\toprule
Benchmark & Condition & Classes & $h$ & $c$ & $V$ & Rate & Acc \\
\midrule
\multirow{5}{*}{\shortstack[l]{MASSIVE\\($n{=}1{,}102$)}}
& Raw-17 $\to$ W5H2 map & 18 & 0.842 & 0.840 & 0.841 & 3.0 & 93.3\% \\
& \textbf{W5H2-8 (ours)} & \textbf{8} & \textbf{0.791} & \textbf{0.777} & \textbf{0.784} & \textbf{3.0} & \textbf{88.8\%} \\
& Target-only (7 cls) & 7 & 0.770 & 0.843 & 0.805 & 2.8 & 94.1\% \\
& Action-only (5 cls) & 5 & 0.491 & 0.686 & 0.573 & 2.3 & 81.8\% \\
& Random-8 (control) & 8 & 0.495 & 0.490 & 0.492 & 3.0 & 68.0\% \\
\midrule
\multirow{3}{*}{\shortstack[l]{BANKING77\\($n{=}3{,}080$)}}
& Raw-77 & 77 & 0.831 & 0.836 & 0.833 & 6.3 & 78.4\% \\
& Grouped-15 & 9 & 0.317 & 0.265 & 0.289 & 3.2 & 47.5\% \\
& Grouped-5 & 5 & 0.255 & 0.215 & 0.233 & 2.3 & 55.2\% \\
\bottomrule
\end{tabular}
\end{table}

\textbf{Key findings.}
(1)~W5H2-8 ($V{=}0.784$) massively outperforms Random-8 ($V{=}0.492$, $+59\%$)---structured decomposition matters, not just grouping to 8 classes.
(2)~Target-only ($V{=}0.805$) slightly outperforms W5H2-8 ($V{=}0.784$), but at the cost of losing action-verb discrimination needed for cache safety (e.g., ``check email'' vs.\ ``send email'' both map to the email domain).
(3)~Action-only ($V{=}0.573$) collapses: precision drops from $h{=}0.791$ to $h{=}0.491$ because 5 action classes cannot separate 8 intent groups.
(4)~Raw-17 with post-hoc mapping achieves the highest V ($0.841$), suggesting that training on fine-grained labels preserves information---but requires $2.25\times$ the training data (144 vs.\ 64 examples).
(5)~BANKING77 grouping ablation shows that \emph{ad hoc} semantic grouping without domain structure collapses catastrophically: Grouped-15 ($V{=}0.289$) and Grouped-5 ($V{=}0.233$) lose both precision and consistency.

The W5H2 decomposition occupies a favorable trade-off: fewer classes than Raw-17 (easier to label, faster to train) while preserving the action--target structure that prevents unsafe cache hits.

\subsection{W5H2 vs.\ Alternative Decompositions}
\label{sec:decomp-comparison}

The ablation in Section~\ref{sec:ablation} varies \emph{granularity} within the W5H2 framework.
A stronger test asks: does the W5H2 (What, Where) decomposition itself outperform principled alternatives?
We compare five decomposition strategies head-to-head (Table~\ref{tab:decomp-compare}), all using SetFit with 8 shots per class and the same backbone (all-MiniLM-L6-v2, seed 42).

\begin{table}[t]
\centering
\caption{Decomposition strategy comparison. All use SetFit 8-shot (seed 42). \emph{Target-only} achieves the highest accuracy on MASSIVE but loses action-verb discrimination critical for cache safety.}
\label{tab:decomp-compare}
\small
\begin{tabular}{@{}lcccccc@{}}
\toprule
Decomposition & \multicolumn{3}{c}{MASSIVE (8-cls)} & \multicolumn{3}{c}{NyayaBench v2 (20-cls)} \\
\cmidrule(lr){2-4} \cmidrule(lr){5-7}
 & Acc.\ (\%) & $V$ & Classes & Acc.\ (\%) & $V$ & Classes \\
\midrule
\textbf{W5H2 (What, Where)} & \textbf{90.1} & \textbf{0.804} & 8 & \textbf{52.4} & \textbf{0.603} & 20 \\
Target-only & 92.7 & 0.819 & 7 & 52.8 & 0.452 & 13 \\
Semantic Role (Action, Patient) & 86.8 & 0.783 & 11 & 49.1 & 0.763 & 49 \\
Nyaya Pramana Ensemble & 82.4 & 0.787 & 8 & 48.0 & 0.516 & 20 \\
Verb-only (ablation) & 74.1 & 0.554 & 5 & 42.8 & 0.641 & 40 \\
\bottomrule
\end{tabular}
\end{table}

\textbf{Key findings.}
(1)~Target-only slightly outperforms W5H2 on MASSIVE accuracy (92.7\% vs.\ 90.1\%) by exploiting the fact that the 8 MASSIVE intents are already well-separated by target domain (email, calendar, weather, etc.).
However, Target-only collapses on V-measure for NyayaBench v2 ($V{=}0.452$ vs.\ $V{=}0.603$), losing the action dimension needed to separate classes within the same domain.
This confirms that \emph{target carries more signal than verb for intent caching}---but the verb component is essential for safety when the domain space is shared across intents.
(2)~Semantic Role labeling (Action, Patient) produces 49 classes on NyayaBench v2---over-sharding that fragments the cache without improving accuracy (49.1\% vs.\ 52.4\%).
(3)~The Nyaya Pramana Ensemble, which routes queries through four epistemological tiers (pratyaksha/direct, anumana/inferential, upamana/analogical, shabda/testimonial), handles 73\% of MASSIVE queries via the upamana+shabda tier alone, but achieves only 82.4\% overall---an 8pp gap behind W5H2.
(4)~Verb-only is consistently the weakest decomposition (74.1\% MASSIVE, 42.8\% NyayaBench), confirming that action verbs alone cannot separate intents.
W5H2's (What, Where) pair provides the best trade-off between accuracy and cache-key discriminability.

\subsection{Hierarchical Canonicalization with Abstention}
\label{sec:hierarchical}

\begin{figure}[t]
\centering
\begin{subfigure}[t]{0.48\textwidth}
\centering
\includegraphics[width=\textwidth]{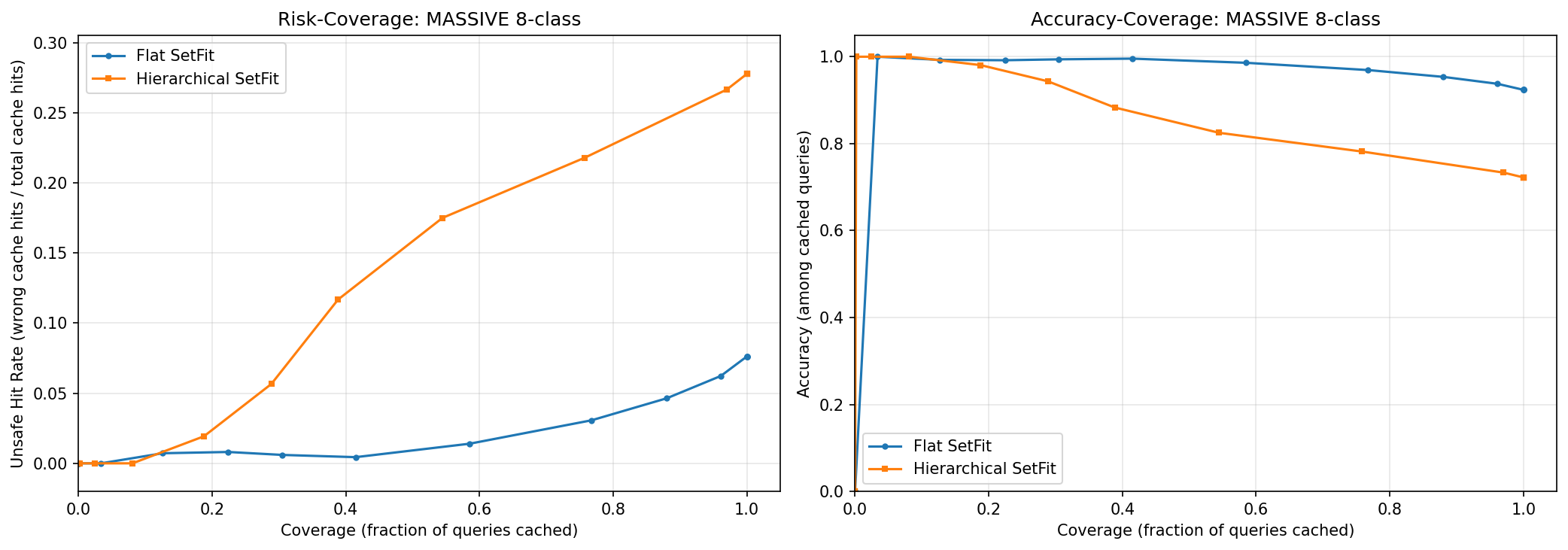}
\caption{MASSIVE (8-class, $n{=}1{,}102$).}
\label{fig:rc-massive}
\end{subfigure}
\hfill
\begin{subfigure}[t]{0.48\textwidth}
\centering
\includegraphics[width=\textwidth]{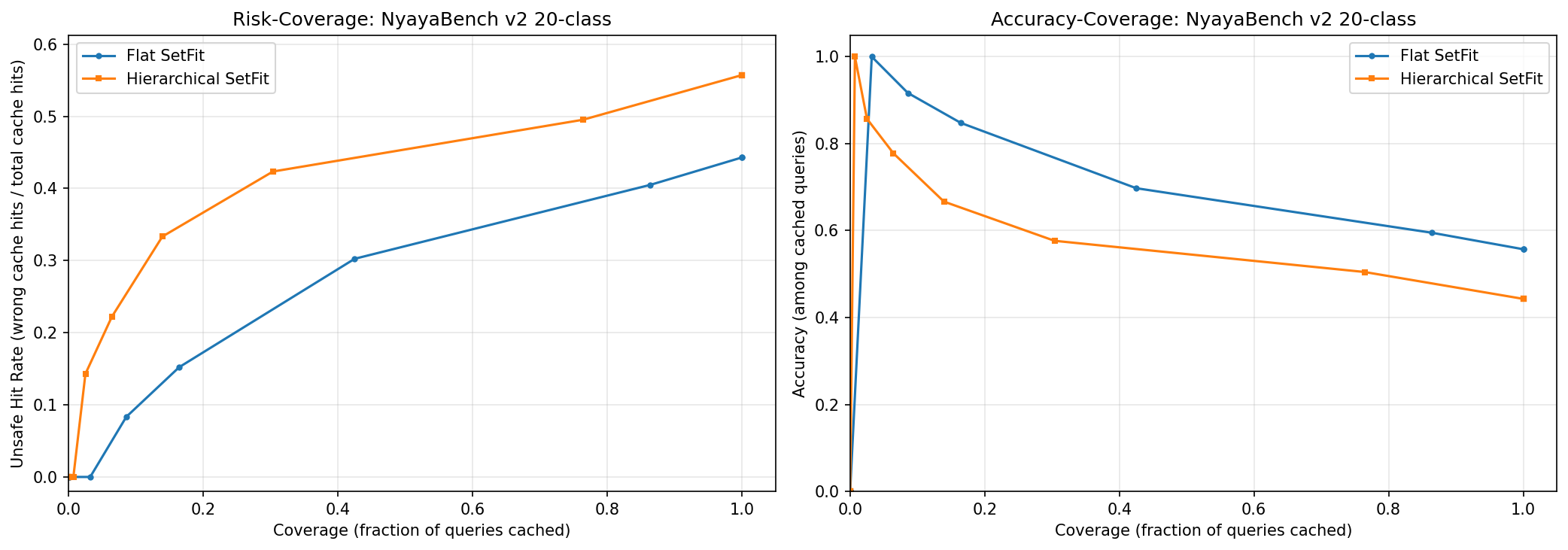}
\caption{NyayaBench v2 (20-class, $n{=}280$).}
\label{fig:rc-nyaya}
\end{subfigure}
\caption{Risk--coverage curves for flat SetFit with abstention.
Higher thresholds $\tau$ reduce coverage but improve accuracy on cached queries.
On MASSIVE, $\tau{=}0.25$ achieves the best safe operating point ($<$5\% unsafe): 88.0\% coverage at 95.4\% accuracy.
On NyayaBench v2, reaching 100\% accuracy requires $\tau{=}0.30$, leaving only 3.2\% coverage---validating the need for the multi-tier cascade.}
\label{fig:risk-coverage}
\end{figure}

The safe canonicalization objective (Eq.~\ref{eq:safe-canon}) requires trading coverage for safety via confidence thresholds.
We evaluate this empirically using flat SetFit with abstention on both MASSIVE and NyayaBench v2, sweeping the threshold $\tau$ from 0 to 0.95 (Figure~\ref{fig:risk-coverage}).

\textbf{MASSIVE} (Table~\ref{tab:risk-coverage}):
At $\tau{=}0$ (no abstention), flat SetFit achieves 92.4\% accuracy at 100\% coverage with 7.6\% unsafe hit rate.
The best safe operating point under a $<$5\% unsafe constraint is $\tau{=}0.25$: 88.0\% coverage, 95.4\% accuracy, 4.6\% unsafe rate.
Raising $\tau$ further to 0.35 yields 98.6\% accuracy at 58.5\% coverage (1.4\% unsafe).
This confirms that flat SetFit on an easy benchmark can safely serve the majority of queries, with the remaining traffic falling to the next cascade tier.

\textbf{NyayaBench v2}:
The 20-class task is fundamentally harder.
At $\tau{=}0$, accuracy is 55.7\% (44.3\% unsafe).
Reaching 100\% accuracy requires $\tau{=}0.30$, but coverage collapses to 3.2\% (only 9 of 280 queries).
Even at $\tau{=}0.20$, coverage is 16.4\% at 84.8\% accuracy.
This validates the cascade design: on hard tasks, SetFit should serve only high-confidence queries and defer the rest.

\begin{table}[t]
\centering
\caption{Selected operating points for flat SetFit with abstention. Full curves in Figure~\ref{fig:risk-coverage}.}
\label{tab:risk-coverage}
\small
\begin{tabular}{@{}llcccc@{}}
\toprule
Benchmark & $\tau$ & Coverage & Acc.\ (cached) & Unsafe \% & $n_\text{cached}$ \\
\midrule
\multirow{4}{*}{MASSIVE}
& 0.00 & 100\% & 92.4\% & 7.6\% & 1,102 \\
& 0.25 & 88.0\% & 95.4\% & 4.6\% & 970 \\
& 0.35 & 58.5\% & 98.6\% & 1.4\% & 645 \\
& 0.60 & 3.4\% & 100\% & 0\% & 37 \\
\midrule
\multirow{4}{*}{\shortstack[l]{NyayaBench\\v2}}
& 0.00 & 100\% & 55.7\% & 44.3\% & 280 \\
& 0.15 & 42.5\% & 69.7\% & 30.3\% & 119 \\
& 0.20 & 16.4\% & 84.8\% & 15.2\% & 46 \\
& 0.30 & 3.2\% & 100\% & 0\% & 9 \\
\bottomrule
\end{tabular}
\end{table}

These results empirically validate the safe canonicalization objective (Eq.~\ref{eq:safe-canon}--\ref{eq:coverage-safety}): the threshold $\tau$ directly implements the coverage--safety tradeoff that the cascade architecture exploits.
On easy tasks (MASSIVE), most traffic can be served from cache; on hard tasks (NyayaBench v2), the cascade correctly defers to more capable tiers.

\subsection{Consistency Regularization}
\label{sec:consistency-reg}

Given that consistency ($c$) is as important as accuracy for cache keys (Section~\ref{sec:consistency}), we investigate whether explicitly regularizing for consistency during training improves SetFit.
We augment each training example with 3 paraphrase variants (word swap, deletion, repetition) and add a consistency loss that penalizes the model when an example and its augmentation receive different predictions.

\begin{table}[t]
\centering
\caption{Vanilla vs.\ consistency-regularized SetFit (5 seeds, 8-shot). Consistency regularization does not improve over baseline SetFit in either regime.}
\label{tab:consist-reg}
\small
\begin{tabular}{@{}llcccc@{}}
\toprule
Benchmark & Method & Accuracy & $V$ & Consistency & Train time \\
\midrule
\multirow{2}{*}{\shortstack[l]{MASSIVE\\(8-class)}}
& Vanilla & 91.56$\pm$1.47\% & 0.819 & 89.4\% & 4.5s \\
& Consist-Reg & 90.92$\pm$1.59\% & 0.808 & 89.0\% & 15.5s \\
& $\Delta$ & $-0.64$ & $-0.011$ & $-0.35$ & $+11.0$s \\
\midrule
\multirow{2}{*}{\shortstack[l]{NyayaBench v2\\(20-class)}}
& Vanilla & 21.6$\pm$1.8\% & 0.179 & 66.5\% & 10.0s \\
& Consist-Reg & 21.7$\pm$1.5\% & 0.188 & 58.8\% & 39.0s \\
& $\Delta$ & $+0.07$ & $+0.009$ & $-7.7$ & $+29.0$s \\
\bottomrule
\end{tabular}
\end{table}

\textbf{Results} (Table~\ref{tab:consist-reg}):
On MASSIVE, consistency regularization marginally \emph{hurts}: accuracy drops from 91.56\% to 90.92\% ($\Delta{=}{-}0.64$pp), V-measure from 0.819 to 0.808, and consistency from 89.4\% to 89.0\%---all within noise.
Training time increases $3.4\times$ (4.5s $\to$ 15.5s).
On NyayaBench v2, both methods achieve ${\sim}$21.6\% accuracy (essentially random among 20 classes), with consistency regularization actually \emph{reducing} consistency (66.5\% $\to$ 58.8\%).

\textbf{Interpretation.}
Baseline SetFit's contrastive training already implicitly optimizes for consistency: the cosine similarity loss encourages all examples of the same class to cluster tightly in embedding space, which naturally produces consistent predictions for paraphrases.
Explicit consistency regularization adds no signal beyond what contrastive learning already captures, while the augmented training data (word swap, deletion) may introduce noisy examples that slightly degrade performance.
This is a useful negative result: practitioners can use vanilla SetFit without additional regularization overhead.

With the accuracy and structure validated, we turn to the practical question: what does this mean for a user's wallet?

\subsection{Cost Model}
\label{sec:cost}

Using real API pricing as of February 2026 (Table~\ref{tab:cost}), we model costs for the five-tier architecture.
Token profiles are based on empirical measurement: a typical agent request uses ${\sim}$1,050 input + ${\sim}$1,200 output tokens; Tier 3 extraction uses ${\sim}$400 input + ${\sim}$100 output; Tier 4 multi-step reasoning uses ${\sim}$2,000 input + ${\sim}$3,000 output.
Our LLM baseline confirms these profiles: 1,102 MASSIVE examples consumed 132,026 input tokens (${\sim}$120/req) and 129,615 output tokens (${\sim}$118/req), totaling \$0.098 at GPT-4o-mini pricing (\$0.089/1K requests) or \$0.091 at DeepSeek V3 pricing.

\begin{table}[t]
\centering
\caption{Monthly cost comparison at 50 requests/day (five-tier architecture). Tier 3: DeepSeek V3.2 (\$0.28/\$0.42 per M tokens). Tier 4: Claude Sonnet 4.5 (\$3/\$15 per M tokens). No-cache baseline uses Claude Sonnet 4.5 for every request. Pricing as of February 2026.}
\label{tab:cost}
\small
\begin{tabular}{@{}lccc@{}}
\toprule
Strategy & Monthly Cost & Savings & Local \% \\
\midrule
No Cache (all Claude Sonnet) & \$31.72 & --- & 0\% \\
APC (6\% hit rate) & \$29.82 & 5.9\% & 6\% \\
GPTCache (37.9\% hit rate) & \$19.70 & 37.9\% & 37.9\% \\
\textbf{W5H2 (85\% local)} & \textbf{\$0.80} & \textbf{97.5\%} & \textbf{85\%} \\
\bottomrule
\end{tabular}
\end{table}

\begin{figure}[t]
\centering
\begin{subfigure}[t]{0.48\textwidth}
\centering
\includegraphics[width=\textwidth]{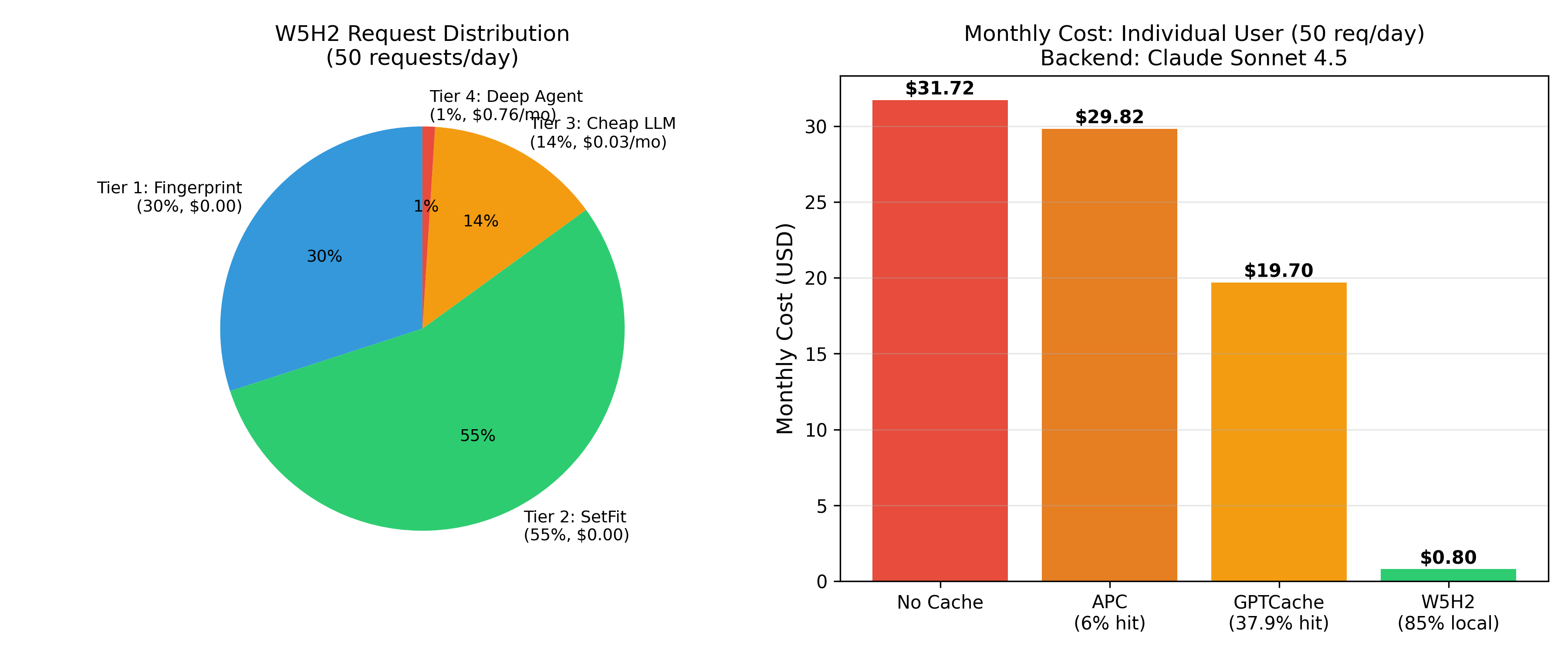}
\caption{Individual user cost breakdown and comparison.}
\label{fig:cost-individual}
\end{subfigure}
\hfill
\begin{subfigure}[t]{0.48\textwidth}
\centering
\includegraphics[width=\textwidth]{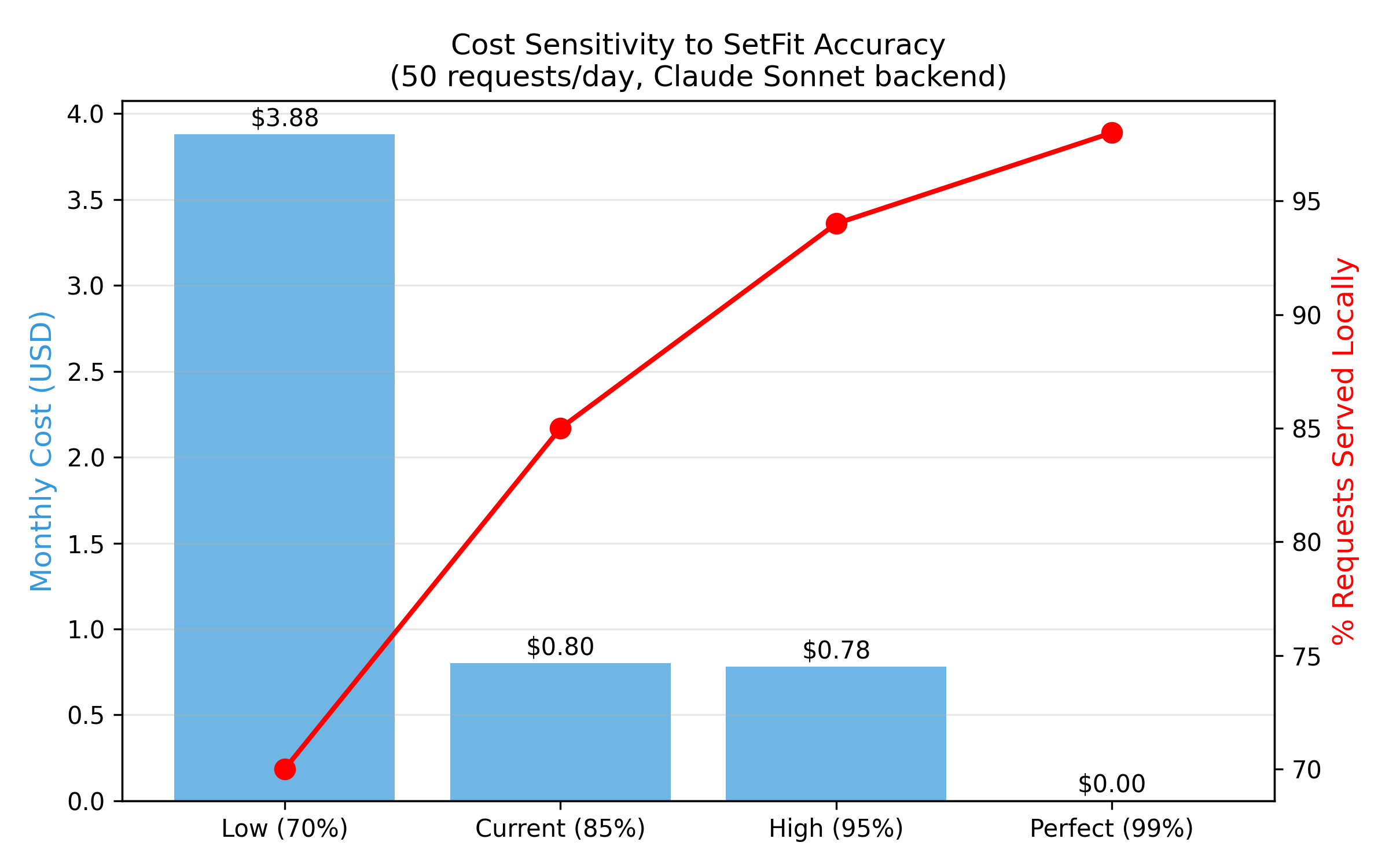}
\caption{Cost sensitivity to SetFit accuracy.}
\label{fig:cost-sensitivity}
\end{subfigure}
\caption{Cost model with real February 2026 API pricing.}
\label{fig:cost}
\end{figure}

The cost reduction holds across scales: startup (50 users, 10K requests/day) pays \$159/month vs.\ \$6,345; enterprise (500 users) pays \$1,595 vs.\ \$63,450.
Sensitivity analysis shows the dominant factor is local tier hit rate, not API pricing: even at ``low'' SetFit accuracy (70\% local), monthly cost is \$3.88---still 87.8\% savings (Figure~\ref{fig:cost}).

One final counter-intuitive result deserves attention: in the cascade's compound-detection stage, the \emph{smaller} model wins.

\subsection{The Over-Entailment Problem}

The compound detection results reveal a general phenomenon: for NLI-based binary classification, model calibration matters more than raw NLI performance.
mDeBERTa produces entailment scores of 0.66--0.96 for \emph{all} inputs against the ``compound query'' hypothesis, making threshold-based classification impossible.
MiniLMv2's weaker but better-calibrated scores enable clean separation.
This has implications beyond caching: any system using NLI models for threshold-based decisions should evaluate calibration, not just accuracy.

These results are encouraging, but we first provide formal guarantees for the threshold selection, then discuss caveats.

\subsection{Risk-Controlled Selective Prediction}
\label{sec:rcps}

The threshold sweeps in Section~\ref{sec:hierarchical} identify good operating points empirically, but provide no statistical guarantee on the unsafe hit rate at deployment.
We apply Risk-Controlling Prediction Sets (RCPS; \citealt{bates2021rcps}) to obtain finite-sample guarantees via Proposition~\ref{prop:rcps}.

\begin{figure}[t]
\centering
\includegraphics[width=\columnwidth]{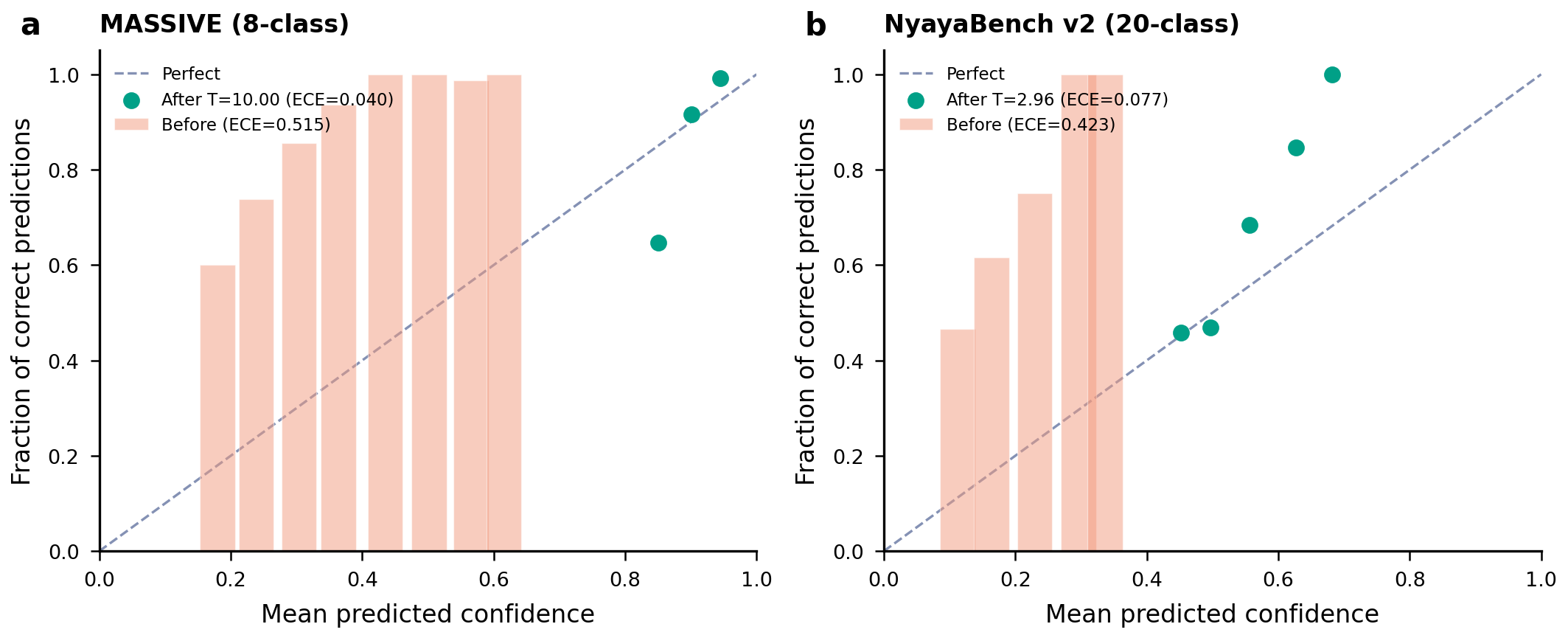}
\caption{Reliability diagrams for SetFit confidence calibration.
\textbf{(a)}~MASSIVE (8-class): moderate calibration; temperature scaling reduces ECE.
\textbf{(b)}~NyayaBench v2 (20-class): higher ECE reflects the harder task; temperature scaling provides larger improvement.
Bars show pre-calibration binned accuracy; dots show post-calibration.}
\label{fig:reliability}
\end{figure}

\textbf{Calibration analysis.}
Before applying RCPS, we assess whether SetFit's confidence scores are well-calibrated using Expected Calibration Error (ECE; \citealt{guo2017calibration}) with 15 bins.
On MASSIVE, ECE$=$0.515, indicating substantial overconfidence---SetFit's softmax outputs cluster near 1.0 even for incorrect predictions.
Temperature scaling (optimizing $T$ on a held-out calibration set via NLL minimization, yielding $T{=}10.0$) reduces ECE to 0.040, a $13\times$ improvement.
On NyayaBench v2, ECE$=$0.423 before and 0.077 after temperature scaling ($T{=}2.97$), a $5.5\times$ improvement.
Figure~\ref{fig:reliability} shows reliability diagrams before and after temperature scaling.

\textbf{RCPS threshold selection.}
We split each benchmark 50/50 into calibration and test sets (stratified, seed 42) and sweep $K{=}100$ thresholds.
For each candidate $\tau$, we compute the UCB under each bound variant (Eq.~\ref{eq:rcps}) and find the minimum $\tau^*$ satisfying the constraint.

\textbf{Ablation of bound variants.}
We compare four finite-sample correction methods: (1)~Hoeffding + union bound (baseline), (2)~Empirical Bernstein + union bound, (3)~LTT fixed-sequence + Hoeffding \citep{angelopoulos2022learn}, and (4)~LTT + Empirical Bernstein.
We additionally evaluate two distributionally robust methods---Wasserstein DRO (at shift budgets $\varepsilon \in \{0.01, 0.05\}$) and CVaR tail-risk bounds---and, for NyayaBench v2, PAC-Bayes bounds with an informative prior transferred from MASSIVE \citep{catoni2007pacbayes}.
All methods achieve \emph{zero} guarantee violations across 18 ($\alpha$, $\delta$) configurations.

On MASSIVE ($n_\text{cal}{=}549$, $\delta{=}0.10$, $\alpha{=}0.10$): LTT eliminates the $\ln K$ union-bound penalty, yielding $\tau^*{=}0.21$ with \textbf{94.0\%} test coverage---a 27\% relative improvement over Hoeffding ($\tau^*{=}0.31$, 73.8\%).
Empirical Bernstein alone provides an intermediate improvement (79.6\% coverage, $\tau^*{=}0.29$).
LTT + Empirical Bernstein matches LTT + Hoeffding at $\alpha{=}0.10$ (both 94.0\%) but separates at tighter risk tolerances: at $\alpha{=}0.02$, LTT + Bernstein achieves 6.0\% coverage where all other methods are infeasible.
Wasserstein DRO and CVaR are strictly more conservative than Hoeffding at the same $\alpha$, since they pay an additional robustness premium; these are appropriate only when distribution shift is a concrete concern.

On NyayaBench v2 ($n_\text{cal}{=}134$, $\delta{=}0.10$, $\alpha{=}0.10$): the Hoeffding term (${\approx}0.161$) exceeds the risk budget, making standard bounds infeasible.
LTT recovers feasibility at $\alpha{=}0.10$ but with only 3.4\% coverage ($\tau^*{=}0.30$).
PAC-Bayes-$\lambda$ with a MASSIVE-transferred prior achieves \textbf{14.4\%} coverage ($\tau^*{=}0.20$) at the same $\alpha{=}0.10$---a $4.2\times$ improvement over LTT---by leveraging the cross-domain risk profile as an informative prior that reduces the effective KL complexity term.
Even without the transfer prior, PAC-Bayes-$\lambda$ achieves 12.3\% coverage, demonstrating that the bound's tighter small-sample behavior outperforms Hoeffding-family methods in this regime.

\begin{figure}[t]
\centering
\includegraphics[width=\columnwidth]{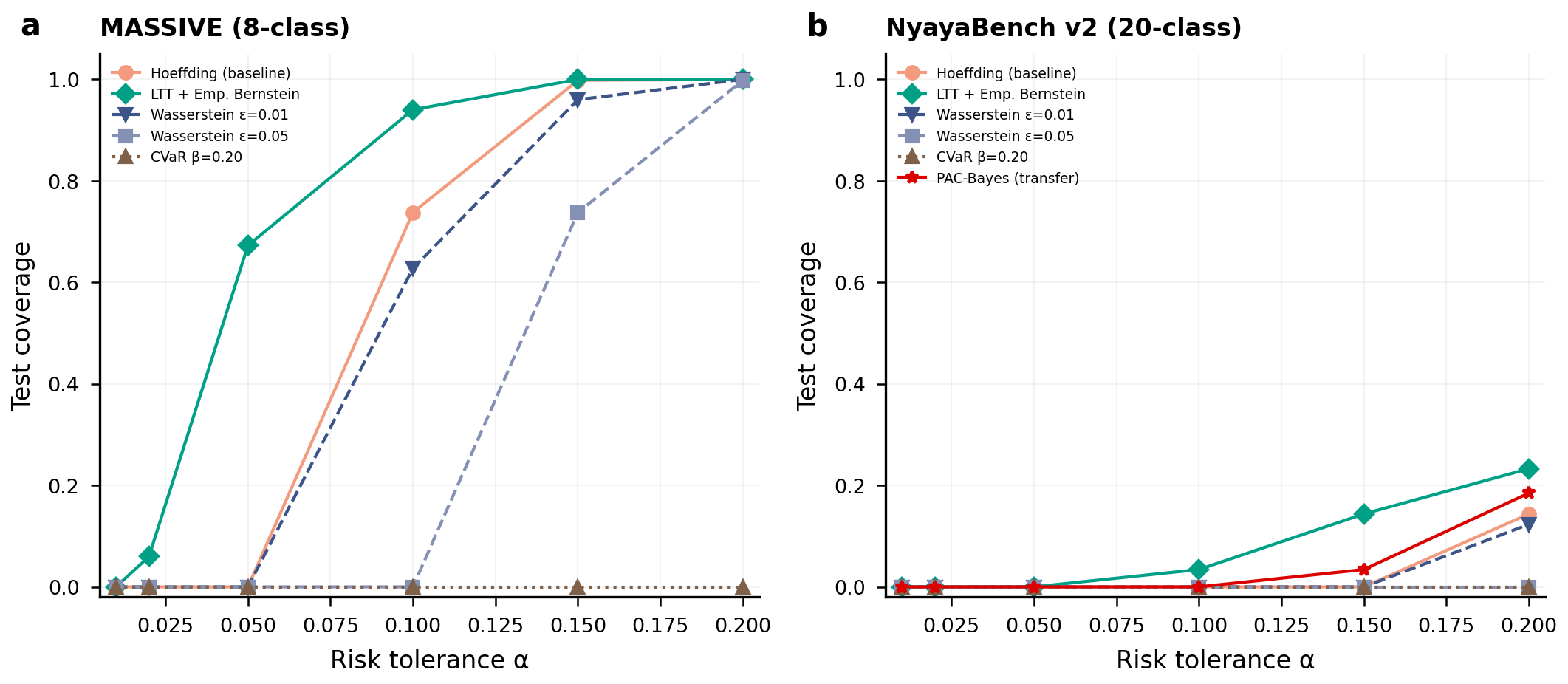}
\caption{Test coverage as a function of risk tolerance $\alpha$ ($\delta{=}0.10$) across bound variants.
\textbf{(a)}~MASSIVE: LTT + Emp.\ Bernstein (green) dominates Hoeffding (orange) across all $\alpha$; Wasserstein DRO (blue) is more conservative by design.
\textbf{(b)}~NyayaBench v2: PAC-Bayes with MASSIVE transfer prior (red) is the only method achieving meaningful coverage at $\alpha{=}0.10$; standard bounds are infeasible below $\alpha{=}0.15$.}
\label{fig:ablation}
\end{figure}

\textbf{Practical implications.}
The ablation reveals a clear recipe: use \emph{LTT + Empirical Bernstein} for large calibration sets ($n \gtrsim 500$), where eliminating the union bound provides the largest gain; switch to \emph{PAC-Bayes with domain transfer} for small calibration sets ($n \lesssim 200$), where informative priors compensate for limited data.
The RCPS framework converts the empirical threshold sweep into a deployment-ready protocol: (1)~collect $n$ labeled calibration examples; (2)~compute the UCB under the appropriate bound; (3)~deploy $\tau^*$ with a certificate that the unsafe rate is bounded by $\alpha$ with probability $1{-}\delta$.
For MASSIVE-scale tasks, LTT + Bernstein achieves 94.0\% guaranteed coverage at $\alpha{=}0.10$, closely matching the empirical operating point ($\tau{=}0.25$, 88\% from Section~\ref{sec:hierarchical}) while providing a formal certificate.
Distributionally robust methods (Wasserstein DRO, CVaR) trade coverage for shift robustness and are best reserved for deployments where the query distribution is expected to evolve.

\subsection{Limitations}
\label{sec:limitations}

\textbf{NyayaBench v2 composition.}
NyayaBench v2 (8,514 entries, 528 intents, 63 languages) is sourced from real agentic usage patterns across voice assistants, smart home devices, IoT controllers, productivity agents, and regional deployments.
The 528 raw intents are mapped to 20 W5H2 super-classes; while the mapping achieves 100\% coverage, some borderline intents (e.g., regional payment methods) could reasonably belong to multiple classes.
We complement NyayaBench v2 with three established benchmarks: MASSIVE (1,102), BANKING77 (3,080), and CLINC150 (4,500).

\textbf{SetFit variance.}
Few-shot learning is sensitive to example selection.
We report mean $\pm$ std across 5 random seeds $\{42, 123, 456, 789, 1024\}$ to quantify this variance.
On MASSIVE (8-class): 8-shot accuracy ranges from 89.1\% to 94.2\% ($\sigma{=}1.7$pp); 16-shot from 90.8\% to 93.8\% ($\sigma{=}1.0$pp)---doubling shots reduces variance by 41\% while increasing mean accuracy by 1.4pp.
On NyayaBench v2 (20-class): 8-shot ranges from 53.6\% to 56.6\% ($\sigma{=}1.0$pp); 16-shot from 60.4\% to 65.0\% ($\sigma{=}1.5$pp)---mean accuracy improves by 7.3pp but variance increases slightly, likely because the larger training set exposes more sensitivity to which examples are drawn from the long-tailed class distribution.

\textbf{Traffic distribution assumptions.}
The five-tier cost model assumes 30\%/40\%/15\%/14\%/1\% traffic distribution across Tiers 0--4 (fingerprint/BERT/SetFit/cheap LLM/deep agent), based on estimated personal agent usage patterns.
These percentages are not empirically measured from production deployments.
Sensitivity analysis (Figure~\ref{fig:cost-sensitivity}) shows cost savings remain substantial ($>$87\%) even at pessimistic SetFit accuracy (70\% local), but real-world distributions may differ.

\textbf{SetFit adversarial robustness.}
Cross-intent confusion examples achieve only 40--50\% accuracy (e.g., ``Send the weather report to John'' $\to$ \texttt{retrieve\_weather}).
This is solvable by adding confusing pairs and retraining (30 seconds), but represents a real limitation.

\textbf{V-measure requires ground truth.}
Production systems lack intent labels.
We envision periodic annotation-based audits rather than continuous monitoring.

\textbf{Supervised baseline gap.}
Fine-tuned BERT with full training data achieves 97.3\% vs.\ SetFit's 91.1\%$\pm$1.7\% with 64 examples.
The ${\sim}$6pp gap is the price of few-shot learning.
The BERT $\to$ SetFit cascade mitigates this: at confidence $>$0.85, BERT handles 97.9\% of queries at 98.2\% accuracy; SetFit catches the remaining 2.1\%.

\textbf{No end-to-end agent deployment.}
Our evaluation measures intent classification and clustering quality as proxies for cache effectiveness.
We do not validate with end-to-end tool execution traces, production replay logs, or user-facing task success rates.
The gap between classification accuracy and actual cache safety (wrong tool invoked due to misclassification) remains unmeasured.
Future work should evaluate on live agent deployments with tool execution correctness as the primary metric.

\textbf{Single-user.}
We evaluate single-user patterns; multi-tenant caching introduces additional challenges around intent namespace collisions.

\section{Conclusion}
\label{sec:conclusion}

Existing agent caching methods fail on personal agent tasks: GPTCache achieves only 37.9\% accuracy on MASSIVE at safe thresholds, APC achieves 0--12\%, and even a 20B-parameter LLM with 4-shot prompting reaches only ${\sim}$79\% at $>$3s latency.
The root cause is that similarity-based and generative methods lack the structured consistency that caching requires.

By reframing agent caching as \emph{canonicalization}---and observing that cache-key evaluation reduces to clustering quality evaluation via V-measure---we connect this problem to databases, compilers, and audio fingerprinting.
The resulting five-tier cascade (fingerprint $\to$ BERT $\to$ SetFit $\to$ cheap LLM $\to$ deep agent) achieves strong results across four benchmarks with just 8 labeled examples per class:
91.1\%$\pm$1.7\% on MASSIVE (8 classes), 85.9\% on CLINC150 (150 classes), 77.9\% on BANKING77 (77 classes), and 55.3\%$\pm$1.0\% on NyayaBench v2 (20 real agentic classes, rising to 62.6\% at 16-shot).
V-measure ranges from 0.504 (NyayaBench v2) to 0.914 (CLINC150), operating at or near the information-theoretic compression rate on established benchmarks.
Cross-lingual transfer reaches 37.7\% across 30 languages with zero non-English training data---a baseline that targeted few-shot examples can improve.

Three additional experiments strengthen the framework.
First, hierarchical canonicalization with abstention (Section~\ref{sec:hierarchical}) empirically validates the safe canonicalization objective: on MASSIVE, a confidence threshold of $\tau{=}0.25$ yields 88\% coverage at 95.4\% accuracy with only 4.6\% unsafe cache hits, while on the harder NyayaBench v2, reaching 100\% accuracy requires $\tau{=}0.30$ at 3.2\% coverage---confirming that the cascade must defer uncertain queries on difficult tasks.
We further strengthen this with finite-sample RCPS guarantees (Section~\ref{sec:rcps}): an ablation of six bound variants shows that LTT fixed-sequence testing \citep{angelopoulos2022learn} with Empirical Bernstein corrections achieves 94.0\% guaranteed coverage on MASSIVE at $\alpha{=}0.10$---a 27\% improvement over Hoeffding---while PAC-Bayes bounds with cross-domain transfer are the only method providing meaningful coverage (14.4\%) on NyayaBench v2's small calibration set.
Second, head-to-head comparison of five decomposition strategies (Section~\ref{sec:decomp-comparison}) shows that W5H2 (What, Where) outperforms the Nyaya Pramana Ensemble by ${\sim}$8pp on MASSIVE and ${\sim}$4.5pp on NyayaBench v2, while the target-only ablation is surprisingly competitive on MASSIVE (92.7\%) but collapses on NyayaBench v2 V-measure ($V{=}0.452$ vs.\ $V{=}0.603$), confirming that the action dimension is essential when domains are shared across intents.
Third, consistency regularization (Section~\ref{sec:consistency-reg}) provides a useful negative result: explicit consistency training does not improve SetFit ($\Delta{=}{-}0.64$pp on MASSIVE, $+0.07$pp on NyayaBench v2) because contrastive learning already implicitly optimizes for consistency---practitioners can use vanilla SetFit without additional overhead.

Under modeled traffic assumptions, the practical projection is \$0.80/month for a 50-request/day personal agent vs.\ \$31.72 for all-LLM---a 97.5\% cost reduction.
Sensitivity analysis shows savings remain above 87\% even under pessimistic accuracy assumptions, though production validation is needed to confirm traffic distribution estimates.


\newpage
\appendix
\section{NyayaBench v2 Per-Language Results}
\label{app:nyayabench}

Table~\ref{tab:perlang} shows SetFit-Multi zero-shot cross-lingual transfer accuracy on NyayaBench v2 across 30 translated languages.
The model is trained on 160 English examples only (8 per class $\times$ 20 W5H2 classes).
Each language has 200 test examples with 200 shared intents mapped to the 20-class taxonomy.

\begin{table}[h]
\centering
\caption{SetFit-Multi per-language accuracy on NyayaBench v2 (20-class, trained on English only). $n{=}200$ per language.}
\label{tab:perlang}
\small
\begin{tabular}{@{}lcl@{}}
\toprule
Language & Accuracy (\%) & Family \\
\midrule
az (Azerbaijani) & 48.0 & Turkic \\
cs (Czech) & 54.0 & Slavic \\
da (Danish) & 42.0 & Germanic \\
el (Greek) & 49.5 & Hellenic \\
fi (Finnish) & 43.5 & Uralic \\
gu (Gujarati) & 40.0 & Indic \\
he (Hebrew) & 42.0 & Semitic \\
hi (Hindi) & 44.5 & Indic \\
hu (Hungarian) & 46.5 & Uralic \\
ka (Georgian) & 49.0 & Kartvelian \\
km (Khmer) & 48.0 & Austroasiatic \\
kn (Kannada) & 20.5 & Dravidian \\
lo (Lao) & 45.5 & Kra-Dai \\
ml (Malayalam) & 13.5 & Dravidian \\
mr (Marathi) & 44.5 & Indic \\
ms (Malay) & 46.5 & Austronesian \\
my (Myanmar) & 11.5 & Sino-Tibetan \\
ne (Nepali) & 29.0 & Indic \\
no (Norwegian) & 50.0 & Germanic \\
pa (Punjabi) & 17.0 & Indic \\
ro (Romanian) & 51.0 & Romance \\
ru (Russian) & 47.5 & Slavic \\
si (Sinhala) & 17.5 & Indic \\
sw (Swahili) & 13.0 & Niger-Congo \\
ta (Tamil) & 12.5 & Dravidian \\
te (Telugu) & 16.5 & Dravidian \\
uk (Ukrainian) & 50.5 & Slavic \\
ur (Urdu) & 42.0 & Indic \\
uz (Uzbek) & 44.0 & Turkic \\
zh (Chinese) & 52.0 & Sino-Tibetan \\
\midrule
\textbf{Mean (30 langs)} & \textbf{37.7} & \\
\bottomrule
\end{tabular}
\end{table}

\section{NyayaBench v2 Per-Class Analysis}
\label{app:pertask}

\begin{figure}[h]
\centering
\includegraphics[width=\columnwidth]{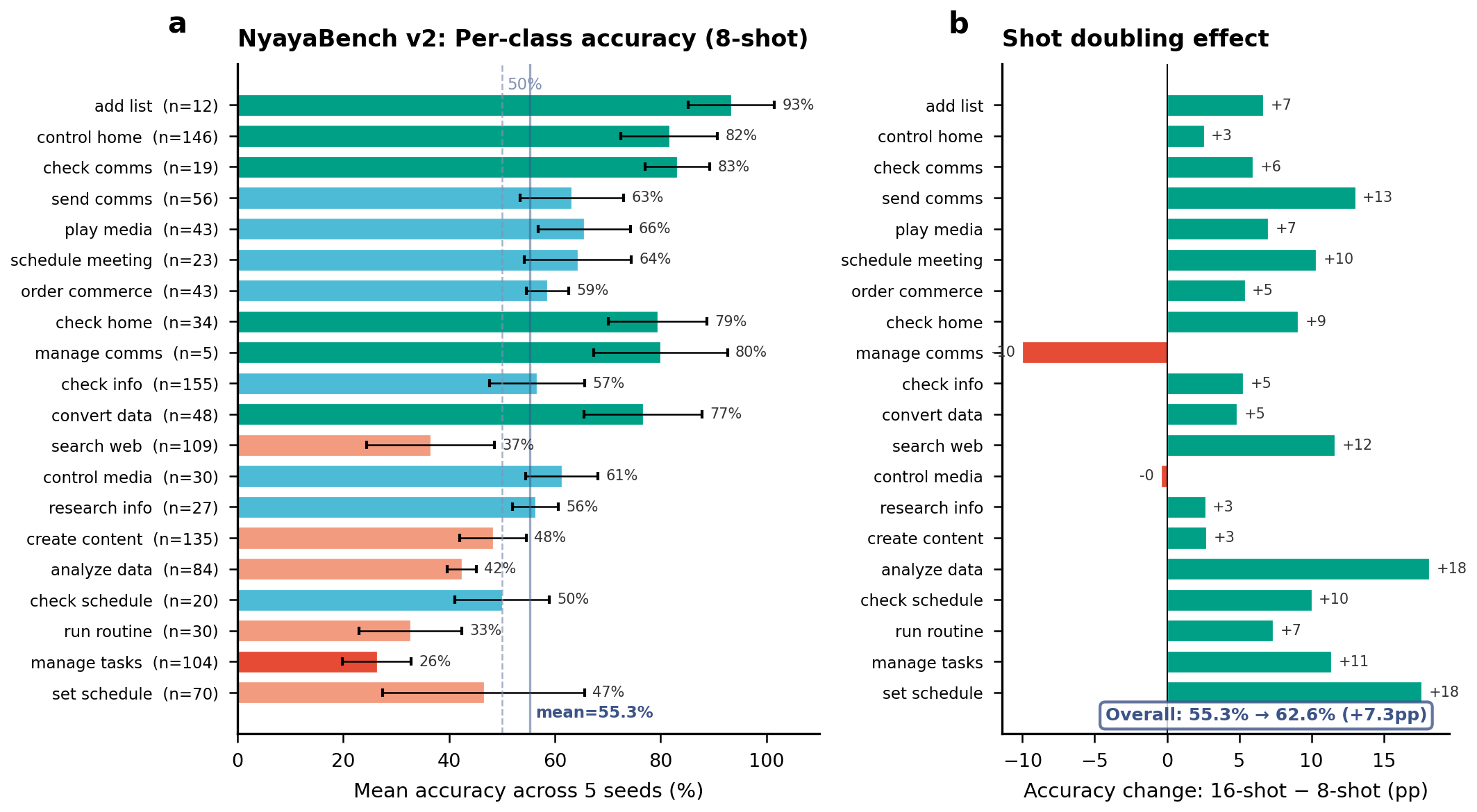}
\caption{NyayaBench v2 per-class analysis.
\textbf{(a)}~SetFit-EN accuracy per W5H2 class (8-shot, 5 seeds), sorted by performance. Error bars show $\pm$1 std across seeds. Color encodes performance tier (green $>$70\%, amber 40--70\%, red $<$40\%).
\textbf{(b)}~Shot-doubling effect: accuracy gain from 8-shot to 16-shot per class. Most classes improve; \texttt{set\_schedule} shows the largest absolute gain.}
\label{fig:perclass}
\end{figure}

Figure~\ref{fig:perclass} visualizes SetFit-EN per-class accuracy on NyayaBench v2 and the effect of doubling training examples.
Table~\ref{tab:pertask} provides the full numerical breakdown.

\begin{table}[h]
\centering
\caption{SetFit-EN per-class accuracy on NyayaBench v2. EN test examples per class vary based on dataset composition.}
\label{tab:pertask}
\small
\begin{tabular}{@{}lcc@{}}
\toprule
W5H2 Class & EN Test $n$ & Accuracy (\%) \\
\midrule
check\_info & 155 & 57.4 \\
control\_home & 146 & 86.3 \\
create\_content & 135 & 48.9 \\
search\_web & 109 & 54.1 \\
manage\_tasks & 104 & 26.9 \\
analyze\_data & 84 & 46.4 \\
set\_schedule & 70 & 15.7 \\
send\_comms & 56 & 73.2 \\
convert\_data & 48 & 56.3 \\
order\_commerce & 43 & 65.1 \\
play\_media & 43 & 72.1 \\
check\_home & 34 & 61.8 \\
run\_routine & 30 & 33.3 \\
control\_media & 30 & 53.3 \\
research\_info & 27 & 51.9 \\
schedule\_meeting & 23 & 65.2 \\
check\_schedule & 20 & 35.0 \\
check\_comms & 19 & 78.9 \\
add\_list & 12 & 100.0 \\
manage\_comms & 5 & 60.0 \\
\bottomrule
\end{tabular}
\end{table}

\section{SetFit Training Details}
\label{app:setfit}

\textbf{Hyperparameters.}
Backbone: \texttt{all-MiniLM-L6-v2} (EN) or \texttt{paraphrase-multilingual-MiniLM-L12-v2} (Multi).
Training: SetFit default contrastive pair generation, cosine similarity loss, 1 epoch body + 1 epoch head.
8 or 16 examples per class.
Seeds: $\{42, 123, 456, 789, 1024\}$; all few-shot results report mean $\pm$ std across 5 seeds.
Training time: 87s (EN, 8 classes) to 5,141s (BANKING77 16-shot, 77 classes).

\textbf{Retraining protocol.}
When Tier 3 produces a new labeled example, it is added to the training pool.
SetFit retrains in ${\sim}$30 seconds (8 classes), enabling continuous improvement without deployment downtime.

\section{Full V-Measure Results}
\label{app:vmeasure}

Table~\ref{tab:vmeasure-full} presents the complete V-measure results including all GPTCache $k$ variants, Majority baseline, and FMI scores.

\begin{table}[h]
\centering
\caption{Full V-measure results across all methods and benchmarks.}
\label{tab:vmeasure-full}
\small
\begin{tabular}{@{}llcccccc@{}}
\toprule
Benchmark & Method & $h$ & $c$ & $V$ & AMI & FMI & Rate (bits) \\
\midrule
\multirow{7}{*}{MASSIVE}
& Oracle & 1.000 & 1.000 & 1.000 & 1.000 & 1.000 & 3.00 \\
& SetFit-EN & 0.817 & 0.810 & 0.813 & 0.810 & 0.833 & 3.00 \\
& GPTCache ($k{=}8$) & 0.784 & 0.788 & 0.786 & 0.784 & --- & 3.00 \\
& LLM (GPT-oss-20b) & 0.638 & 0.662 & 0.650 & 0.645 & --- & 3.00 \\
& GPTCache ($k{=}4$) & 0.534 & 0.835 & 0.652 & 0.650 & --- & 2.00 \\
& Majority (1 key) & 0.000 & 1.000 & 0.000 & 0.000 & 0.408 & 0.00 \\
& Random / APC & 1.000 & 0.277 & 0.433 & 0.000 & 0.000 & 10.11 \\
\midrule
\multirow{5}{*}{BANKING77}
& Oracle & 1.000 & 1.000 & 1.000 & 1.000 & 1.000 & 6.27 \\
& SetFit 16-shot & 0.857 & 0.862 & 0.860 & 0.820 & --- & 6.27 \\
& SetFit 8-shot & 0.840 & 0.845 & 0.843 & 0.798 & 0.660 & 6.27 \\
& GPTCache ($k{=}77$) & 0.788 & 0.808 & 0.798 & 0.743 & --- & 6.27 \\
& GPTCache ($k{=}30$) & 0.621 & 0.824 & 0.708 & 0.673 & --- & 4.91 \\
\midrule
\multirow{4}{*}{CLINC150}
& Oracle & 1.000 & 1.000 & 1.000 & 1.000 & 1.000 & 7.23 \\
& SetFit 8-shot & 0.911 & 0.916 & 0.914 & 0.868 & --- & 7.23 \\
& GPTCache ($k{=}150$) & 0.887 & 0.902 & 0.894 & 0.841 & --- & 7.23 \\
& GPTCache ($k{=}30$) & 0.592 & 0.894 & 0.713 & 0.670 & --- & 4.91 \\
\bottomrule
\end{tabular}
\end{table}

\section{Cost Model Details}
\label{app:cost}

\textbf{API Pricing (February 2026).}
Tier 3: DeepSeek V3.2 (\$0.28 input / \$0.42 output per M tokens).
Tier 4: Claude Sonnet 4.5 (\$3.00 input / \$15.00 output per M tokens).
All-LLM baseline uses Claude Sonnet 4.5 for every request.

\textbf{Token profiles.}
Full agent request: 50 (query) + 500 (system prompt) + 200 (context) + 300 (tool schemas) = 1,050 input tokens; 800 (plan) + 400 (formatting) = 1,200 output tokens.
Tier 3 extraction: 400 input + 100 output.
Tier 4 deep agent: 2,000 input + 3,000 output.

\textbf{Scaling analysis.}
\begin{table}[h]
\centering
\caption{Monthly cost scaling across deployment sizes.}
\label{tab:scaling}
\small
\begin{tabular}{@{}lrrrrc@{}}
\toprule
Scale & Req/day & No Cache & GPTCache & W5H2 & Savings \\
\midrule
Individual & 50 & \$31.72 & \$19.70 & \$0.80 & 97.5\% \\
Power User & 200 & \$126.90 & \$78.80 & \$3.19 & 97.5\% \\
Small Team (5) & 1,000 & \$634.50 & \$394.02 & \$15.95 & 97.5\% \\
Startup (50) & 10,000 & \$6,345 & \$3,940 & \$159 & 97.5\% \\
Enterprise (500) & 100,000 & \$63,450 & \$39,402 & \$1,595 & 97.5\% \\
\bottomrule
\end{tabular}
\end{table}

\end{document}